\documentclass{article}

\usepackage[nonatbib,preprint]{neurips}

\usepackage[numbers]{natbib}

\usepackage{makecell}
\usepackage{amsmath}
\usepackage[utf8]{inputenc}
\usepackage[T1]{fontenc}  
\usepackage{xcolor}
\usepackage{adjustbox}
\usepackage{enumitem}
\usepackage{tocloft}

\usepackage[
  colorlinks=true,
  linkcolor=gray!40!black, 
  citecolor=blue!40!black,  
  urlcolor=blue!40!black 
]{hyperref}

\usepackage{url}          
\usepackage{booktabs}   
\usepackage{amsfonts}     
\usepackage{nicefrac}   
\usepackage{pifont}
\usepackage{tablefootnote} 
\usepackage{graphicx}
\usepackage{xspace}
\usepackage{wrapfig}
\usepackage{amsmath}
\usepackage{multirow}
\usepackage{subcaption}

\usepackage[capitalize]{cleveref}
\crefname{section}{Sec.}{Secs.}
\Crefname{section}{Section}{Sections}
\crefname{table}{Tab.}{Tabs.}
\Crefname{table}{Table}{Tables}
\crefname{figure}{Fig.}{Figs.}
\Crefname{figure}{Figure}{Figures}
\crefname{equation}{Eq.}{Eqs.}
\Crefname{equation}{Equation}{Equations}
\usepackage{amsfonts,amssymb}

\usepackage[utf8]{inputenc}
\usepackage{amsmath}
\usepackage{algorithm}
\usepackage{algorithmic}

\usepackage{caption}
\usepackage{setspace}

\usepackage{tocloft}
\usepackage{tabularx}
\usepackage{xcolor}
\usepackage[table,xcdraw]{xcolor}
\usepackage{wrapfig}
\usepackage{titletoc}
\usepackage[section]{placeins}

\definecolor{myblue}{RGB}{0,102,204}   
\definecolor{mypurple}{RGB}{123, 31, 162}

\usepackage{booktabs}
\usepackage{array}

\usepackage{colortbl}
\newcommand{\nocolorbox}[2]{%
  \begingroup
  \setlength{\fboxsep}{0pt}
  \colorbox{#1}{#2}%
  \endgroup
}

\definecolor{bestcolor}{HTML}{77B776} 
\definecolor{secondcolor}{HTML}{B1D7B1} 
\definecolor{thirdcolor}{HTML}{D8EBD8} 

\newcommand{\best}{\cellcolor{bestcolor}\bfseries}
\newcommand{\second}{\cellcolor{secondcolor}\bfseries}
\newcommand{\third}{\cellcolor{thirdcolor}}

\makeatletter
\DeclareRobustCommand\onedot{\futurelet\@let@token\@onedot}
\def\@onedot{\ifx\@let@token.\else.\null\fi\xspace}

\def\Ours{MetricAnything\xspace}

\title{\Ours: Scaling Metric Depth Pretraining \\with Noisy Heterogeneous Sources}

\author{%
\textbf{Baorui Ma}\thanks{Equal contribution.}\thanks{Corresponding author. \texttt{Correspondence to \{mabaorui2014@gmail.com\}.}},
\textbf{Jiahui Yang}\footnotemark[1],
\textbf{Donglin Di}\thanks{Project leader.},
\textbf{Xuancheng Zhang},\\
\textbf{Jianxun Cui},
\textbf{Hao Li},
\textbf{Xie Yan},
\textbf{Wei Chen} \\[3.5pt]
Li Auto Inc
}

\begin{document}

\maketitle

\vspace{-2.8em} 

\begin{abstract}
Scaling has powered recent advances in vision foundation models, yet extending this paradigm to metric depth estimation remains challenging due to heterogeneous sensor noise, camera-dependent biases, and metric ambiguity in noisy cross-source 3D data.
We introduce \textbf{Metric Anything}, a simple and scalable pretraining framework that learns metric depth from noisy, diverse 3D sources without manually engineered prompts, camera-specific modeling, or task-specific architectures. Central to our approach is the \textbf{Sparse Metric Prompt}, created by randomly masking depth maps, which serves as a universal interface that decouples spatial reasoning from sensor and camera biases.
Using $\sim$20M image–depth pairs spanning reconstructed, captured, and rendered 3D data across 10,000+ camera models, we demonstrate—for the first time—a clear scaling trend in the metric depth track. The pretrained model excels at prompt-driven tasks such as depth completion, super-resolution and Radar-camera fusion, while its distilled prompt-free student achieves state-of-the-art results on monocular depth estimation, camera intrinsics recovery, single/multi-view metric 3D reconstruction, and VLA planning. We also show that using pretrained ViT of Metric Anything as a visual encoder significantly boosts Multimodal Large Language Model capabilities in spatial intelligence.
These results show that metric depth estimation can benefit from the same scaling laws that drive modern foundation models, establishing a new path toward scalable and efficient real-world metric perception. We open-source MetricAnything at \url{https://metric-anything.github.io/metric-anything-io/} to support community research.

\end{abstract}

\par\vspace{-11pt}
\noindent\begin{minipage}{\linewidth}
    \centering
    {\setlength{\abovecaptionskip}{2pt}\setlength{\belowcaptionskip}{2pt}%
    \includegraphics[width=0.85\linewidth]{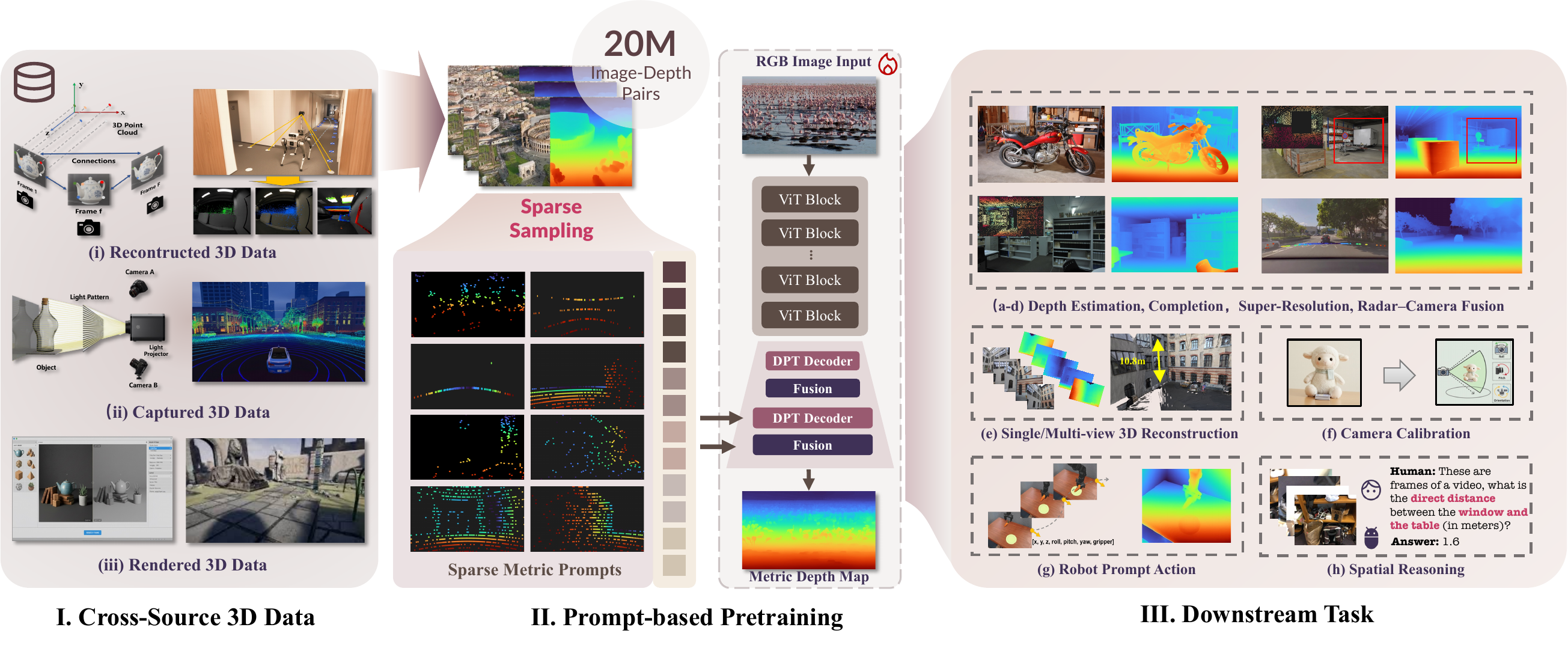}
    \captionof{figure}{\footnotesize \textbf{Overview of Metric Anything.}
    (\uppercase\expandafter{\romannumeral 1}) We aggregate diverse open-source 3D data into per-pixel metric depth maps, forming a $\sim$20M image–depth dataset captured by over 10,000 cameras across heterogeneous scenes.
    (\uppercase\expandafter{\romannumeral 2}) Sparse Metric Prompts, generated by randomly masking depth maps, provide a minimal interface that decouples spatial reasoning from sensor and camera biases, enabling metric depth learning from noisy, heterogeneous sources.
    (\uppercase\expandafter{\romannumeral 3}) The pretrained model and its distilled prompt-free student generalize robustly across multiple downstream tasks, revealing a clear scaling trend and establishing a solid foundation for versatile, data-driven metric perception.}
    \label{fig:pipeline}}
\end{minipage}
\par\vspace{-4pt}

\newpage
\setlength{\cftbeforesecskip}{8pt}      
\setlength{\cftbeforesubsecskip}{6pt}    
\setlength{\cftbeforesubsubsecskip}{4pt} 
\begin{spacing}{0.9}
\tableofcontents
\end{spacing}
\newpage

\vspace{-6mm}
\section{Introduction}
Vision foundation models have achieved remarkable progress through scaling—training larger models on ever-expanding datasets to unlock emergent capabilities and robust generalization~\cite{dosovitskiy2020vit,oquab2023dinov2,kirillov2023segment,radford2021learning,ramesh2021zero}. As AI systems increasingly interact with the physical world through robotics, AR, and autonomous driving, foundation models must go beyond 2D perception to perceive the 3D world. Monocular depth estimation (MDE) serves as a fundamental bridge, providing depth cues essential for physical interaction.

\begin{figure}[t]
    \centering
    \begin{subfigure}[t]{0.48\linewidth}
        \centering
        \includegraphics[width=\linewidth]{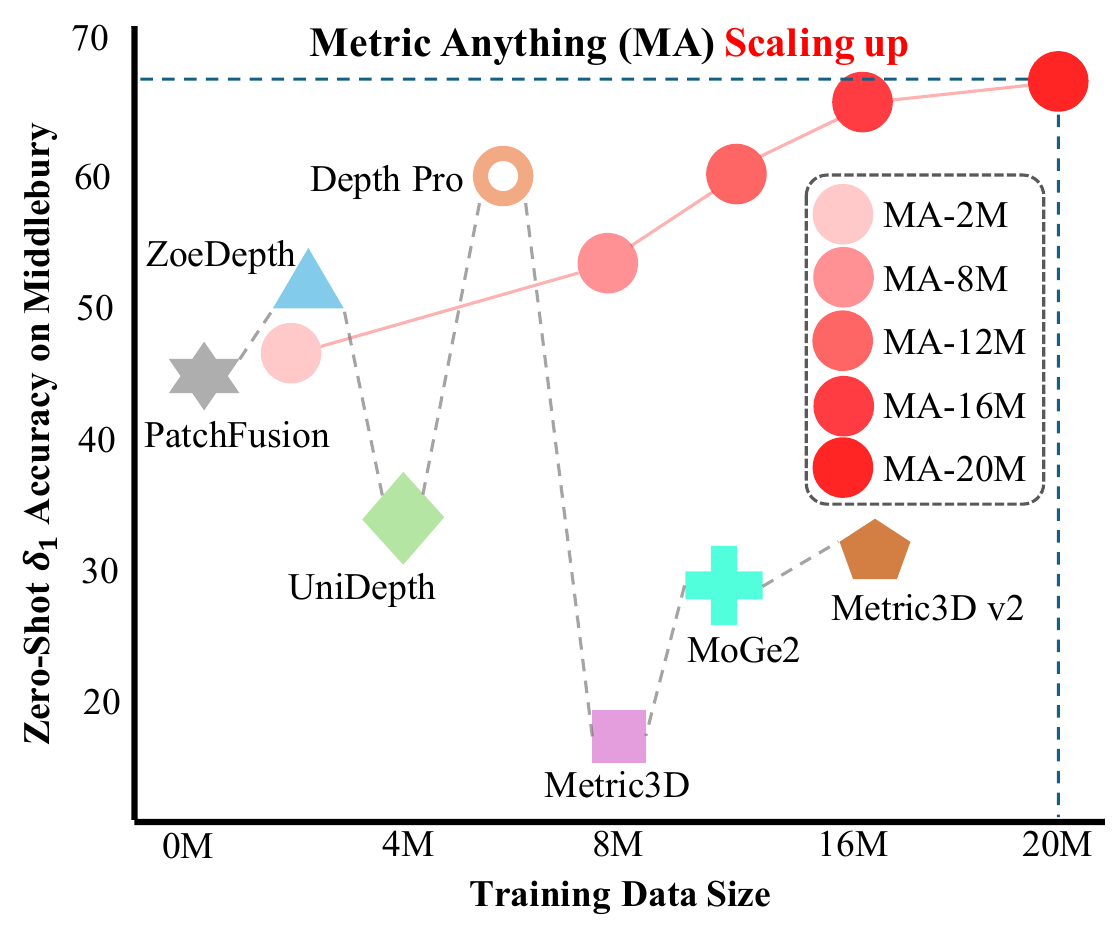}
        \caption{\textbf{Scaling trend.} Larger training dataset yields consistently higher zero-shot $\delta_{1}$ accuracy.}
        \label{fig:scaleup}
    \end{subfigure}
    \hfill
    \vspace{2pt}
    \begin{subfigure}[t]{0.48\linewidth}
        \centering
        \includegraphics[width=\linewidth]{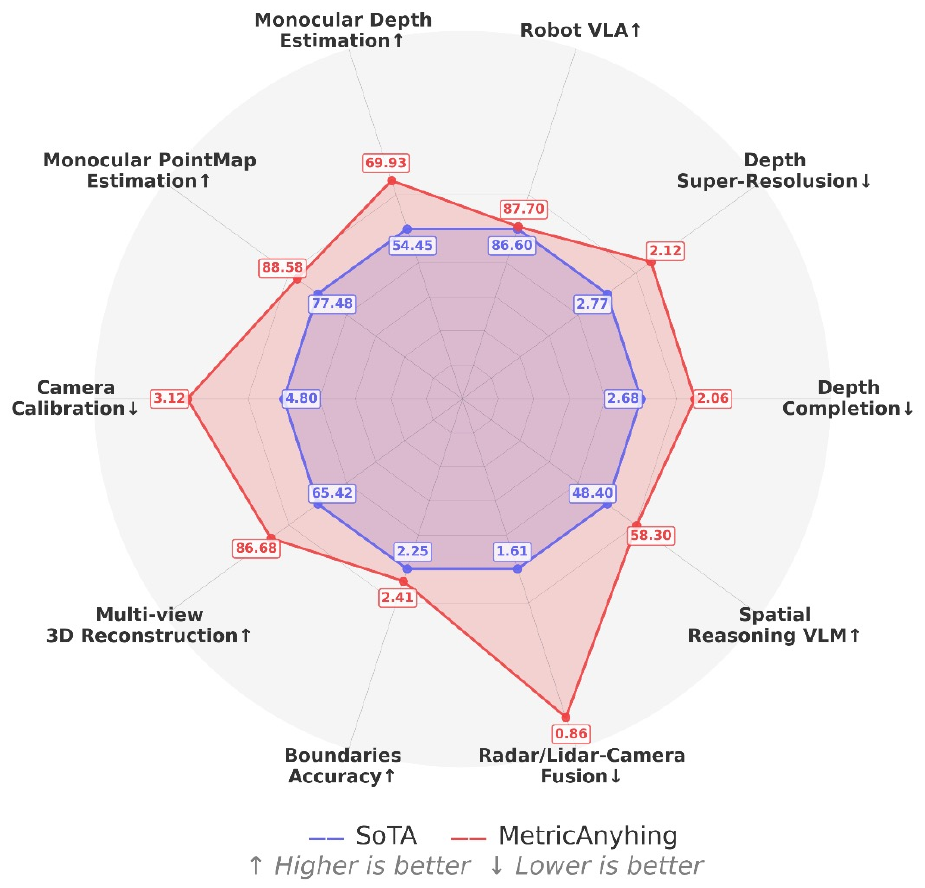}
        \caption{\textbf{Task performance.} Radar plot over downstream tasks; Larger area indicates better performance.}
        \label{fig:radar}
    \end{subfigure}
    \caption{\textbf{Scaling and Generalization.} \Ours exhibits a clear scaling trend and strong overall downstream performance.}
    \label{fig:intro_scale_radar}
\end{figure}

While relative depth estimation has demonstrated successful scaling through synthetic data and large-scale pseudo-label distillation~\cite{depth_anything_v1,depth_anything_v2}, \textbf{metric depth estimation has not exhibited similar scaling trends in previous works}, as illustrated in Fig.~\ref{fig:scaleup}. Unlike relative depth's ordinal relationships, metric depth requires learning absolute, physically meaningful distances—a fundamentally harder problem compounded by data-related challenges.

The core bottleneck lies in the heterogeneity of metric depth data sources. Real-world 3D annotations come from diverse sensors (LiDAR, RGB-D, stereo cameras) or reconstruction algorithms (SfM, MVS, SLAM), each introducing distinct noise patterns, hardware-specific artifacts, and camera-dependent biases. This heterogeneity creates three critical issues: (i) significant domain gaps between sources, (ii) noisy supervision from sensor misalignment and algorithmic matching failures, and (iii) metric ambiguity due to varying camera intrinsics. Together, these factors prevent the ``data soup'' scaling strategy that has proven successful in other domains, limiting current metric depth methods to careful curation of small, clean datasets.

Recent approaches have attempted to address these challenges through prompt-based methods \cite{lin2024promptda,wang2025depthprior,zuo2024omni,liu2024depthlab,viola2024marigolddc}, using sparse depth points or simulated LiDAR cues to guide depth prediction. However, these methods remain limited in scope—they typically focus on specific  prompt-driven downstream tasks with small datasets and rely on complex, hand-crafted prompt construction pipelines that introduce strong human priors. This task-specific engineering limits both scalability and generalization.

\textbf{In this work, we present a fundamentally different perspective}: rather than engineering better prompts for specific tasks, we demonstrate that a simple, scalable prompt-based pretraining paradigm can unlock the metric depth perception capabilities with large-scale heterogeneous 3D data.  In other words, we are not aiming to construct a  stronger model for a particular prompt-driven task (e.g., depth completion or super-resolution). We posit a simple yet effective pre-training paradigm to demonstrate a new direction, which proves that with effective mitigation of data scarcity issues, scaling trends can similarly occur in the metric estimation track, just as in NLP and 2D vision. 
Our key insight is that sparse prompts can serve as a universal interface to decouple spatial understanding from sensor-specific biases, enabling effective learning from diverse, noisy sources without complex engineering.


To realize this vision, we introduce \textbf{Metric Anything}, a minimalist pretraining paradigm tailored for metric depth estimation, which generates \textbf{Sparse Metric Prompts} by randomly masking portions of depth maps. This design intentionally avoids task-specific architectures, complex prompt construction rules, or strong 3D inductive biases, making specialized network design unnecessary. Instead, it enables the model to naturally learn structural and metric understanding directly from data. To support scaling, we assemble approximately 20M image--depth pairs across three categories: reconstructed 3D data (SfM/SLAM/MVS), captured 3D data (LiDAR/ToF/RGB-D), and rendered 3D data. All samples are aligned with metric annotations; for raw point clouds, we obtain per-pixel depths by projection using known calibration. The collection spans over 10{,}000 camera models and diverse environments, enabling unified pretraining across heterogeneous 3D sources.


The pretrained model, with its simple architecture and pretext pretraining objective, excels in prompt-based tasks such as depth completion, super-resolution, and radar-camera depth estimation. Leveraging a dedicated distillation process, we further create a prompt-free student model that achieves state-of-the-art performance across diverse downstream tasks, including monocular depth estimation, camera calibration, multi-view metric 3D reconstruction, and Vision-Language-Action (VLA) planning. We further demonstrate that leveraging the pretrained ViT from our Metric Anything as a visual encoder substantially enhances the spatial reasoning capabilities of Multimodal Large Language Models. Remarkably, without any task-specific design, both the pretrained and distilled models consistently achieve state-of-the-art performance across these tasks.

Our findings align with ``The Bitter Lesson''~\cite{sutton2019bitter}: general-purpose, data-driven methods systematically outperform hand-crafted designs. By demonstrating consistent performance gains with increasing scale (see Fig.~\ref{fig:scaleup})  and robust generalization (Fig.~\ref{fig:radar}) across ten downstream tasks, we establish prompt-based pretraining as a scalable path toward general-purpose metric depth perception. Our contributions are:

\vspace{2pt}
\begin{itemize}
    \item[$\bullet$] \textbf{Metric Anything}: A minimalist prompt-based pretraining paradigm that employs  Sparse Metric Prompt to decouple spatial understanding from sensor biases, enabling scalable learning from heterogeneous 3D sources.
        
    \item[$\bullet$] \textbf{Demonstrated Scaling}: Aggregation of 20M diverse depth-image pairs reveals stable scaling trends in metric depth estimation, previously unseen in this domain.
    \item[$\bullet$] \textbf{Universal Generalization}: Both pretrained and distilled models achieve state-of-the-art performance across ten downstream tasks without task-specific engineering.
\end{itemize}

\section{Related Work}

\subsection{Monocular Depth Estimation}

\textbf{Relative depth estimation.} Over the past decade, relative depth has emerged as the dominant formulation because it can learn from heterogeneous and weakly supervised signals—pairwise orderings \cite{chen2016single,chen2020oasis}, stereo/MVS cues \cite{godard2019digging,li2018megadepth}, and pseudo-depth \cite{ranftl2020midas,depth_anything_v1}—yielding broad data coverage and strong cross-domain generalization. Early CNNs treated depth as regression \cite{eigen2014depth,eigen2015predicting,laina2016deeper}. Subsequent work introduced adaptive discretizations and per-pixel classification \cite{fu2018deep,bhat2021adabins,bhat2022localbins} to better handle non-uniform distributions and ambiguity, but these methods often coupled models to dataset-specific depth ranges, limiting transfer. A key advance was large-scale, multi-dataset training with scale-and-shift-invariant objectives: MegaDepth \cite{li2018megadepth} highlighted the benefits of diverse training data, MiDaS \cite{ranftl2020midas} established mixing heterogeneous datasets with scale-and-shift-invariant losses, and transformer-based architectures brought global context beyond CNN locality constraints \cite{dosovitskiy2020vit,ranftl2021dpt}. 
Depth Anything further demonstrated a practical scaling route for relative depth by combining synthetic supervision that is  noise-free and highly consistent with massive pseudo-labell distillation (62 million images), producing strong generalization \cite{depth_anything_v1,depth_anything_v2}. Replicating this success for metric depth remains an open challenge. Prior works \cite{hu2024metric3d,yin2023metric3d,bhat2023zoedepth,depthpro,wang2025moge2,piccinelli2024unidepth,piccinelli2025unidepthv2} have recognized that metric-aware perception is inherently more challenging than relative depth perception and requires substantially more data for effective learning. 
However, metric depth data, originating from diverse algorithms or hardware configurations, exhibits various types of noise and heterogeneous patterns, further impeding the benefits obtainable from scaling. 
Unlike prior approaches, our prompt-based pretraining paradigm explicitly addresses this inherent noise and heterogeneity, establishing a viable scaling path for robust metric depth estimation.



\noindent \textbf{Metric Depth Estimation.}
Metric depth estimation has progressed from domain-constrained, fixed-intrinsics settings with limited RGB-D/LiDAR supervision \citep{eigen2014depth, yin2019enforcing, guizilini2023towards} to camera-aware, open-domain formulations that couple depth with intrinsics or canonical camera transforms \citep{bhat2023zoedepth, yin2023metric3d, hu2024metric3d, piccinelli2024unidepth, depthpro, wang2025moge2}. Representative strategies include attaching domain-specific metric heads to relative-depth backbones (ZoeDepth) \citep{bhat2023zoedepth}, learning canonical camera normalization for zero-shot transfer (Metric3D and its scaled variants) \citep{yin2023metric3d, hu2024metric3d}, predicting canonical inverse depth with high-resolution ViTs (DepthPro) \citep{depthpro}, and directly regressing metric 3D points with camera-conditioned features (UniDepth) \citep{piccinelli2024unidepth}. MoGe-2 further decouples metric scale from affine-invariant point maps \citep{wang2025moge2, wang2024moge}. While MoGe-2 seeks gains via complex, non-learnable pipelines (e.g., edge detection and depth refinement) to exploit large-scale real data. In this work, we discard hand-crafted processing and propose a pretraining paradigm from a data-driven perspective that  learns a robust representation of metric depth from noisy and heterogeneous 3D sources, enabling the model to convert coarse data into a scalable source of accurate pseudo-labels.



\subsection{Metric Depth Estimation with Sparse Priors}

This line of work predicts dense metric depth from RGB with sparse cues such as LiDAR, RGB-D, or radar \cite{park2024depth,zhang2023completionformer,zuo2024ogni,cheng2019learning,cheng2020cspn++,tang2024bilateral}. Representative approaches include: Marigold-DC, which repurposes a diffusion model for guided denoising with dynamic scale and shift to achieve zero-shot densification \cite{viola2024marigolddc}; OMNI-DC, which predicts multi-scale depth gradients and integrates them to reduce long-range drift, augmented with a Laplacian loss for ambiguous regions and scale normalization for cross-dataset transfer \cite{zuo2024omni}; and methods like PriorDA, which align metric priors with relative depth and refine structure under metric and geometric constraints using pixel-level alignment and distance-aware reweighting \cite{wang2025depthprior}. Other efforts explore foundation model prompting, such as PromptDA, which injects low-resolution noisy depth via a multi-scale fusion decoder but relies on heavy data pipelines involving LiDAR noise simulation, 3D reconstruction hole filling, and mixed supervision \cite{lin2024promptda}. Sensor-specific designs have also emerged, e.g., TacoDepth, which fuses radar and camera data in a single stage using graph-based radar structure extraction and pyramid fusion \cite{wang2025tacodepth}. Despite these advances, most methods remain confined to a single prompt type (e.g., for depth completion or super-resolution) or a single sensor modality. While PriorDA attempts to handle more diverse prompts, it still focuses on hand-designed prompt simulation—such as creating missing areas, downsampling GT, or adding outliers and blur—and remains limited to small-scale training datasets. In contrast, our work shows that minimizing human prior design in prompts and instead learning from naturally heterogeneous data offers a more general and scalable approach. Through simple pretext tasks and data scaling, our model learns essential and robust spatial understanding, achieving state-of-the-art performance across diverse downstream tasks—including multi-sensor fusion, monocular depth estimation, camera calibration, multi-view 3D reconstruction, and VLA policy—without being confined to any single sub-task.

\section{Method}

We aim to develop a general-purpose MDE foundation model that achieves both fine structural fidelity and accurate metric perception, enabling robust generalization across diverse tasks.
To this end, we first aggregate a 20M-scale image–depth dataset (Sec.~\ref{sec:data}) from open-source 3D datasets. We introduce a sparse metric prompt-based pretraining paradigm that supports scalable learning from heterogeneous sources (Sec.~\ref{sec:pretrain}). Finally, the pretrained knowledge is distilled into a prompt-free student model for  broader applicability (Sec.~\ref{sec:student}). An overview is shown in Fig.~\ref{fig:pipeline}.

\subsection{Multi-Source Data Collection} \label{sec:data}

To enable scalable pretraining, we aggregate open-source 3D datasets with metric annotations and standardize all inputs as metric depth maps with validity masks:
\begin{equation}
\{\mathbf{G}, \mathbf{M}\}, \quad \mathbf{G} \in \mathbb{R}^{H \times W}, \;\; \mathbf{M} \in \{0,1\}^{H \times W},
\end{equation}
where $\mathbf{G}$ denotes the metric depth along the camera $z$-axis, and $\mathbf{M}$ indicates valid measurements ($\mathbf{M}(p)=1$ for valid pixels and $\mathbf{M}(p)=0$ otherwise).

\vspace{2pt}
\noindent \textbf{Reconstructed 3D  Data.}
We collect depth maps from open-source datasets \cite{baruch2021arkitscenes,huang2018deepmvs-mvs-synth,tosi2021unrealstereo4k,wen2025stereo-fsd,yang2019drivingstereo,bauer2019uasol,cordts2016cityscapes,wang2020gta-sfm} produced by a wide range of reconstruction techniques, such as SfM, SLAM, MVS, binocular/multi-view stereo (e.g., plane-sweeping, PatchMatch), and dense temporal stereo. These depth maps are represented as per-pixel metric depth $\mathbf{G}$ with validity mask $\mathbf{M}$, as defined above. Such maps often contain artifacts and missing regions, especially in weakly textured areas, specular or metallic surfaces, thin structures, sharp boundaries, repetitive patterns, depth discontinuities, occlusions, or under motion blur and dynamic scenes. These errors typically arise from incorrect or lost matches during reconstruction.

\vspace{2pt}
\noindent \textbf{Captured 3D Data.}
We also incorporate point clouds captured by LiDAR, ToF, and RGB-D sensors \cite{gehrig2021dsec,xiao2021pandaset,houston2020one-lyft,zamir2018taskonomy,ramakrishnan2021hm3d,wilson2023argoverse,Matterport3D,straub2019replica,sun2020waymo,antequera2020mapillary-psd,guizilini2020ddad,dai2017scannet}. Given the camera intrinsics $\mathbf{K}$ and sensor-to-camera pose $\mathbf{T}_{c \leftarrow s}=[\mathbf{R}\mid \mathbf{t}]$, depth maps $\{\mathbf{G}, \mathbf{M}\}$ are obtained by projecting each sensor-frame point $\mathbf{X}_j^s$ onto the corresponding camera plane.
Sensor-derived depth is often noisy and sparse due to limitations in sensor resolution and the influence of environmental factors such as severe weather. Additional sources of error include beam divergence and sampling gaps (affecting thin structures), weak or anomalous returns on specular, metallic, or transparent/semi-transparent surfaces, multi-path interference, mixed pixels at sharp boundaries, out-of-range distances, occlusions, and motion-induced time offsets. These factors result in invalid measurements, indicated by $\mathbf{M}(p)=0$ for affected pixels.


\vspace{2pt}
\noindent \textbf{Rendered 3D Data.}
To complement the real-world data, we also include a small portion of depth maps rendered from virtual engines \cite{wrenninge2018synscapes,fonder2019mid-air,roberts2021hypersim,gaidon2016vkitti,wang2019irs,wang2020tartanair,apollosynthetic,li2023matrixcity,gomez2025all-urbansyn,zheng2020structured3d,gomez2025all-urbansyn,ros2016synthia}.  
Although these synthetic depth maps exhibit limited visual realism and diversity, they are completely noise-free, preserve fine structural details, and display sharp geometry, thereby providing valuable geometric supervision.

\vspace{2pt}
In total, this purposefully aggregated dataset comprises approximately 20 million image–depth pairs, captured by over 10,000 distinct camera models and spanning a wide variety of real and synthetic scenes.

\subsection{Pretraining via Sparse Metric Prompt}  \label{sec:pretrain}



\begin{figure}[t]
    \centering
    \includegraphics[width=0.8\linewidth]{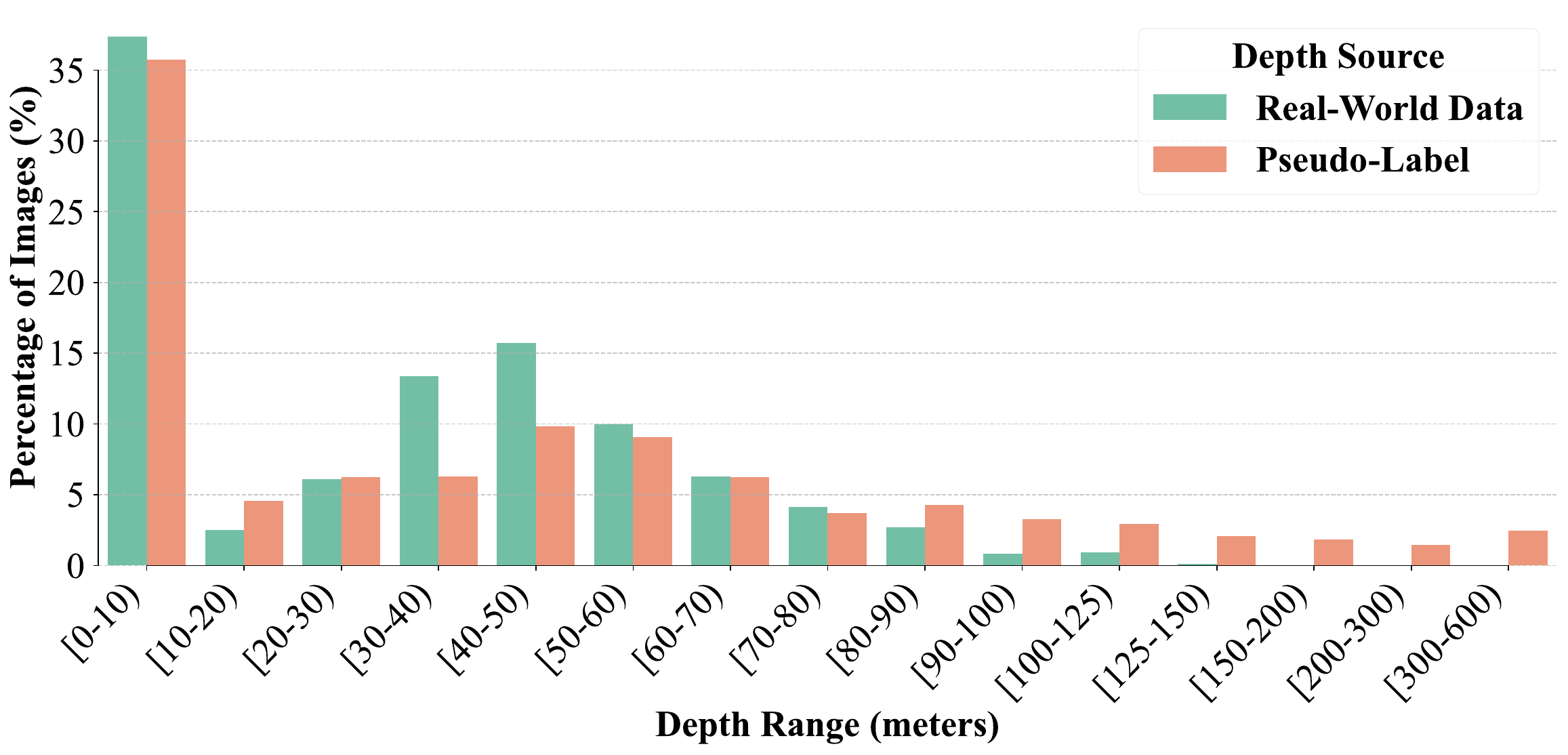}
    \caption{Percentile Depth Range Comparison from Seven Datasets (Real-world  vs. Our Pseudo Labels).}
    \label{fig:depthrange}
\end{figure}

\paragraph{Problem Formulation.}

Given a monocular image $I \in \mathbb{R}^{H \times W \times 3}$ and a corresponding sparse metric prompt, 
\begin{equation}
P = \{(x_i, y_i, d_i)\}_{i=1}^{N}, 
\end{equation}
where each triplet $(x_i, y_i, d_i)$ represents a pixel coordinate $(x_i, y_i)$ and its associated metric depth value $d_i$. 
The objective of pretraining is to learn a function $f_{\theta}$, parameterized by $\theta$, that predicts a dense metric depth map $D$ from the input image $I$, conditioned on the prompt $P$:
\begin{equation}
D = f_{\theta}(I, P), \quad D \in \mathbb{R}^{H \times W}.
\end{equation}
Here, prompt $P$ provides metric depth for a small subset of pixels, offering limited geometric cues.
The model $f_{\theta}$ learns to propagate these sparse metric constraints across the image, yielding a complete, geometrically consistent, and metrically accurate depth map. 
Since our data sources inevitably contain inherent noise and varying patterns of incompleteness,  $f_{\theta}$ is trained to identify and correct potential errors within the sparse prompt $P$.

\paragraph{Prompt Preparation.}  
From each source in $\{\mathbf{G}, \mathbf{M}\}$, we randomly sample \(N \in [2{,}000, 40{,}000]\) valid pixels (around 1\% of all image pixels) from its depth map to construct a sparse metric prompt \(P\) and its corresponding mask \(M'\).  
Due to diverse patterns, sparsity, and incompleteness across $\{\mathbf{G}\}$, the sparsely sampled prompt \(P\) inevitably inherits heterogeneous distributions and irregular structures, which challenge models to effectively handle diverse distributions.

To mitigate this, we first apply  a prompt preparation step that maps all prompts into a shared intermediate domain.  
Following PriorDA~\cite{wang2025depthprior}, we apply a pre-trained depth prediction model~\cite{depthpro} to obtain a prior depth map \(P_d\).  
Then, inspired by See3D~\cite{ma2025you}, we perform Pixel-wise Depth Scale Alignment (PDSA) and Global Metric Depth Recovery (GMDR) to fill the missing regions in \(P\) under the guidance of \(P_d\).  
The final input prompt is constructed by concatenating the PDSA-refined map, the GMDR-corrected map, and the prompt mask \(M'\) along the channel dimension. This yields a unified and regularized representation of shape $H \times W \times 3$, avoiding issues with irregular prompt network design and training efficiency, while also easing the handling of diverse prompts.




\paragraph{Prompt Injection.}
Although this preparation process can maintain metric consistency between $P_d$ and Prompt $P$ through pixel-wise and global alignment, its parameter-free (non-learning) nature remains sensitive to sampling patterns and noise propagated from prompt $P$. To further address this, we employ a prompt-injection mechanism that allows the model to correct noisy prompts and generate accurate, dense depth predictions.
Following prior works, several strategies can be used for prompt injection, including adaptive layer normalization (AdaLN)~\cite{peebles2023scalable-dit}, cross-attention~\cite{vaswani2017attention,rombach2022high}, ControlNet-style conditioning~\cite{zhang2023adding-controlnet}, conditioned DPT heads~\cite{lin2024promptda}, and conditioning layers parallel to the RGB input~\cite{wang2025depthprior}. 
Among these, we opt for a conditioned DPT head, as it introduces only about 5\% extra parameters while maintaining efficiency through lightweight interpolation and shallow convolutional layers.


This design allocates most parameters to the foundational depth backbone, allowing the main network pathway to correct noise and structural inconsistencies, rather than relying on the injected condition layers. Importantly, we make no bespoke modifications to the backbone; instead, we  reuse a general, efficient, and widely validated architecture~\cite{depthpro}. The only difference is that we merge the patch encoder and image encoder into a single shared ViT to achieve a balance between accuracy and efficiency, as detailed in our ablation studies. This emphasizes our deliberate shift away from hand-crafted priors and manually designed prompts, relying entirely on data-driven learning. By isolating conditioning from architectural changes, we can cleanly assess the effectiveness of the learned prompts and rigorously study the scaling behavior.

\paragraph{Pre-training Objective.}


Following DepthPro~\cite{depthpro},  
we use mean absolute error (MAE) and scale-and-shift invariant mean absolute gradient error (SSI-MAGE) losses for synthetic data. For real-world data, 
we adopt a robust MAE loss that drops top-$n$-largest-loss regions per image during training to mitigate noise sensitivity, where $n$ is set as 20$\%$. The overall training objective is formulated as:
\begin{equation} \label{eq:loss}
L_{total} = \alpha  L_{MAE}  + \beta L_{SSI-MAGE}.
\end{equation}

\begin{figure}[t]
    \centering
    \includegraphics[width=0.8\linewidth]{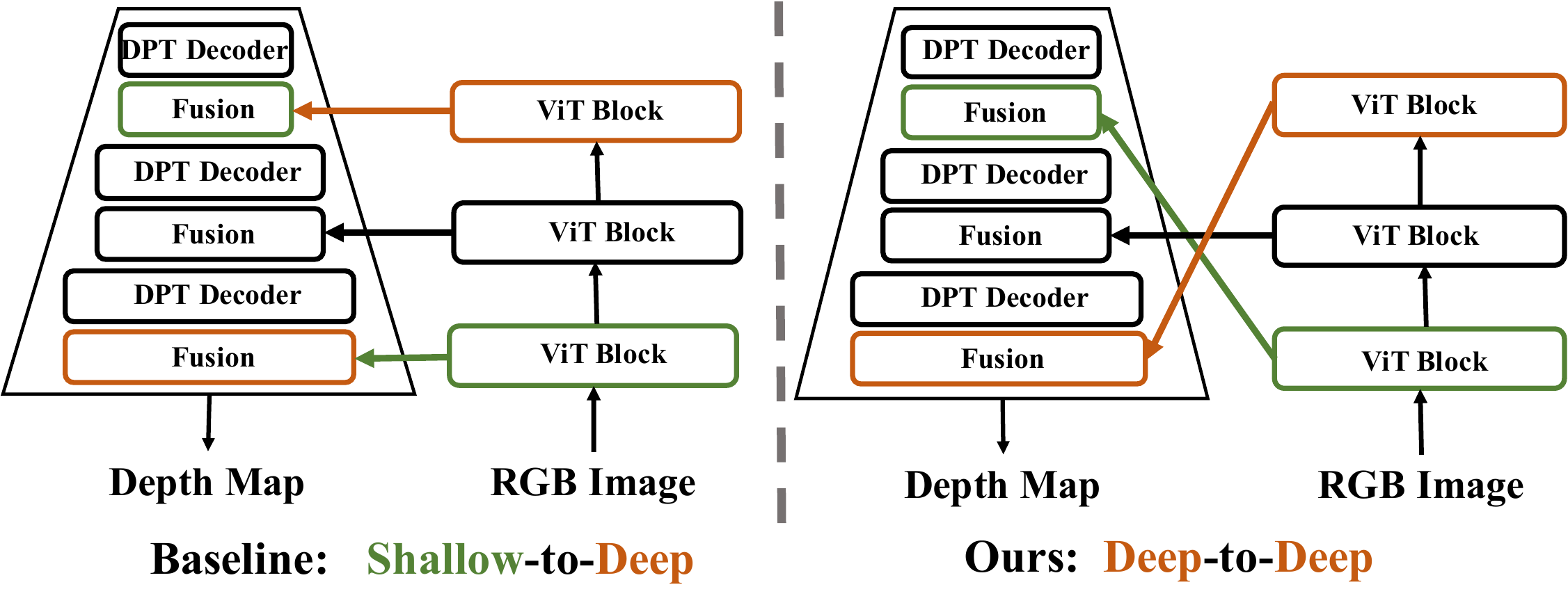}
    \caption{Skip-Connection in ViT-DPT Architecture.}
    \label{fig:skip}
\end{figure}

\subsection{Prompt-Free Model Distillation} \label{sec:student}

\paragraph{Distillation Process.}


To validate the benefits of large-scale pretraining and transfer them to scenarios where prompts are unavailable, we distill the pre-trained (teacher) model into a prompt-free student that performs dense depth prediction solely from RGB inputs.

\vspace{2pt}
Specifically, we leverage the  trained teacher model to generate high-fidelity pseudo depth maps conditioned on sparse prompt $P$ for all real-world images in the collected dataset via a simple feed-forward process, effectively transforming sparse and noisy prompts into dense, high-quality supervision signals (i.e., pseudo labels).
Unlike conventional RGB-D or LiDAR sensors—whose effective sensing range is constrained by hardware to only a few to tens of meters—and many depth-estimation algorithms that struggle to estimate distant backgrounds, our teacher-generated pseudo labels span diverse environments and cover both near and far distances.
These teacher labels train a prompt-free monocular depth estimation model, preserving the teacher’s metric understanding without prompts.

\paragraph{Student Selection.}
A straightforward approach is to directly reuse the pre-training network architecture as the student backbone while removing the prompt layers. However, in our preliminary experiments, this approach resulted in suboptimal performance when trained on the teacher labels, despite many prior works \cite{depthpro,wang2025moge2,hu2024metric3d,bhat2023zoedepth} successfully training similar ViT–DPT configurations from scratch. 

\vspace{2pt}

We hypothesize two main causes for the suboptimality. 
First, our teacher model generates reliable pseudo labels across both near and far regions, unlike most real-world datasets biased toward short-range depths (see Fig.~\ref{fig:depthrange} for a comparison). This wider depth distribution exposes limitations of conventional depth losses: direct depth supervision (L1/L2) blurs fine geometric details, while inverse-depth loss decays too rapidly with distance, losing effective supervision for distant regions. Consequently, these standard losses are suboptimal for supervising our high-fidelity, large-range pseudo labels.
Second,  typical ViT encoders with DPT-head decoders use U-Net-style skip connections, injecting shallow features into deeper layers and propagating deep features upward. While this stabilizes training under noisy supervision, since the low-level cues in the ViT encoder (e.g., textures and colors) are more consistent and easier to learn. The low-level feature is connected via a skip connection to the DPT head near the output, which mitigates conflicts between the output and the noisy supervision, thereby smoothing gradient fluctuations. However, it can underutilize high-level semantic cues essential for precise depth. In our setting, pseudo labels are generated by a unified model, exhibit minimal domain gap, and have been refined to correct most noisy. As a result, we can reduce reliance on shallow-to-deep feature injection and explore more aggressive network designs that fully exploit the rich semantic cues provided by deep block of the ViT encoder.

\paragraph{Improved Student.}
Drawing on these observations and analyses, we retain the multi-scale fusion mechanism proposed by \cite{depthpro} while introducing two key modifications. 
First, we design a distance-balanced inverse-depth loss that preserves fine-grained sensitivity in  near regions while extending effective supervision to long-distance areas. Depth values in log-space are  defined as:

\begin{equation}\label{eq:ours_log_distance}
D_{log} = 1-\ln(x) / {\ln(C)}, 
\end{equation}
where $C$ is a hyperparameter that controls the trade-off between long-range and short-range supervision.

Second, leveraging the high-fidelity teacher labels, we invert the conventional skip-connection scheme between the ViT encoder and DPT head: injecting deep, high-level ViT features into the deeper decoding layers near the output, while shallow, low-level features feed into the shallower decoder layers, as shown in Fig.~\ref{fig:skip}.
This inversion emphasizes semantic reasoning at the final prediction stage,  fully exploiting the rich semantic cues embedded in the teacher-generated pseudo labels, which already exhibit low noise and high structural consistency.
In this way, we experimentally demonstrate that  the student model achieves stable, prompt-free metric depth estimation, effectively distilling the metric perception capability from the pre-trained model.

\section{Experiment}
\label{sec:experiment}

\begin{figure}
    \centering
    \includegraphics[width=1\linewidth]{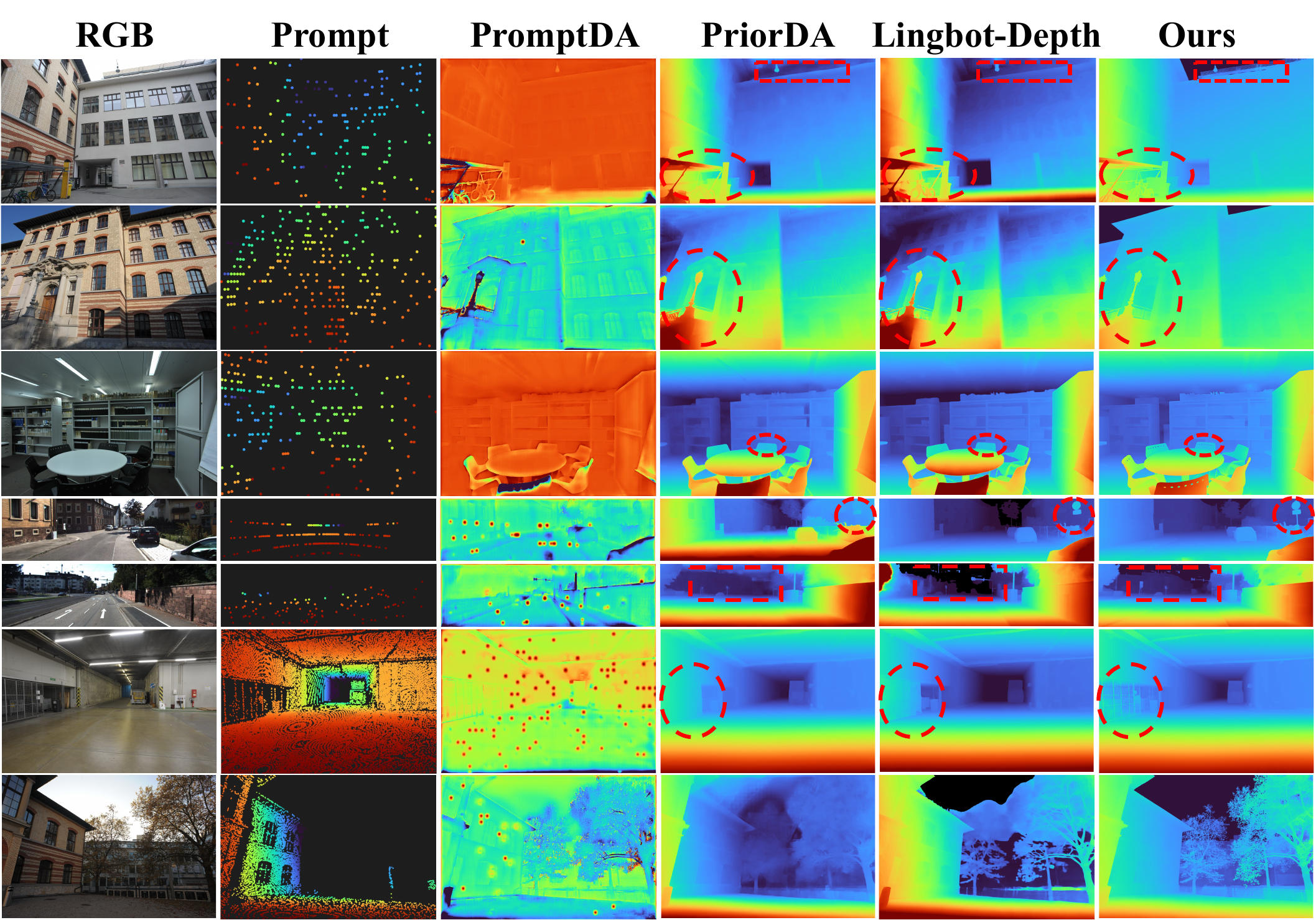}
    \caption{\textbf{Visualization of Depth SR and Completion.}  Our method better recovers missing regions with improved structure. }
\label{fig:teacher_vs_anno}
\end{figure}

\subsection{Prompt-Based Downstream Task}

\subsubsection{Zero-Shot Depth Super-Resolution and Completion.} 

We study depth super-resolution and completion in a zero-shot setting, where our pretrained model directly takes sparse or low-resolution depth maps as prompts, without any task-specific finetuning.
We compare two categories of baselines: (a) post-aligned MDE, including DepthAnything V2~\cite{depth_anything_v2} and DepthPro~\cite{depthpro}; and (b) prior-based MDE, including LingBot-Depth~\cite{lingbot-depth2026}, ~\cite{lin2024promptda}, PriorDA~\cite{wang2025depthprior}, DepthLab~\cite{liu2024depthlab}, Omni-DC~\cite{zuo2024omni}, and Marigold-DC~\cite{viola2024marigolddc}.
Following PriorDA~\cite{wang2025depthprior}, we construct four prompt types—LiDAR-like sparse scans, extremely sparse samples (100 points), and depth maps downsampled by 8× and 16×.
All prompts are directly fed into our pretrained model for zero-shot inference on unseen datasets, including NYUv2~\cite{silberman2012indoor_nyuv2}, ETH3D~\cite{schoeps2017cvpr_eth3d}, and KITTI~\cite{geiger2012_kitti}, covering indoor, outdoor, and open-world scenarios.

As shown in Tab.~\ref{tab:zeroshot_completion} and Fig.~\ref{fig:teacher_vs_anno}, our pretrained model demonstrates strong zero-shot performance across all prompt types and datasets, consistently outperforming both post-aligned and prior-based baselines. Unlike previous approaches \cite{wang2025depthprior,zuo2024omni} that generate synthetic prompts (e.g., LiDAR simulation or noisy downsampling) to approximate test-time conditions, our model is trained only once with simple, sparsely sampled prompts and operates fully zero-shot manner, without any task-specific design or prompt alignment. This enables superior generalization across diverse prompt densities, spatial layouts, and scene domains.

\begin{table}[h]
\centering
\footnotesize
\caption{\footnotesize  Zero-shot Depth Super-Resolution / Completion (AbsRel \%).}
\label{tab:zeroshot_completion}
\resizebox{\columnwidth}{!}{
\begin{tabular}{lcccccccccccc}
\toprule
& \multicolumn{4}{c}{NYUv2 $\downarrow$}
& \multicolumn{4}{c}{ETH-3D $\downarrow$}
& \multicolumn{4}{c}{KITTI $\downarrow$} \\
\cmidrule(lr){2-5} \cmidrule(lr){6-9} \cmidrule(lr){10-13}

Method & $8\times$ & $16\times$ & LiDAR & Extreme  
& $8\times$ & $16\times$ & LiDAR & Extreme  
& $8\times$ & $16\times$ & LiDAR & Extreme \\
\midrule

DAv2 \cite{depth_anything_v2}
& 4.77 & 5.13 & 4.85 & 4.77
& 6.27 & 7.38 & 7.41 & 6.51
& 9.54 & 11.22 & 8.86 & 9.25 \\

DepthPro \cite{depthpro}
& 4.48 & 4.83 & 4.47 & 4.41
& 5.88 & 6.79 & 5.31 & 6.51
& 6.76 & 9.16 & 6.05 & 6.19 \\

Omni-DC \cite{zuo2024omni}
& \second 1.57 & 3.11 & 2.12 & 2.63
& 1.86 & 4.09 & \second 1.88 & 1.98
& 4.05 & 8.35 & 5.27 & 4.17 \\

Marigold-DC
& 1.83 & 3.32 & \second 1.90 & 2.13
& 2.33 & 4.75 & 2.27 & 2.03
& 5.17 & 9.47 & 6.88 & 5.62 \\

DepthLab \cite{liu2024depthlab}
& 2.60 & 3.73 & 4.30 & 6.30
& 2.60 & 4.50 & 6.40 & 8.01
& 17.17 & 22.90 & 37.17 & 40.29 \\

PromptDA \cite{lin2024promptda}
& 1.61 & \best 1.75 & 17.59 & 16.96
& 1.80 & 2.56 & 18.86 & 18.18
& \second 3.92 & \second 4.95 & 21.96 & 21.39 \\

PriorDA \cite{wang2025depthprior}
& 1.73 & 2.79 & 2.01 & \best 2.01
& 2.06 & 3.91 & 1.90 & \second 1.61
& 4.54 & 8.20 & \second 4.81 & 3.76 \\

LingBot-Depth~\cite{lingbot-depth2026}
& 2.23 & 3.05 & 2.19 & \second 2.03
& \best 1.18 & \best 1.54 & 13.19 & 4.12
& 4.24 & 7.13 & 6.55 & \best 3.26 \\

\midrule

\textbf{Ours-Pretrain}
& \best 1.53 & \second 1.86 & \best 1.70 & 2.08
& \second 1.46 & \second 2.03 & \best 0.87 & \best 0.84
& \best 2.34 & \best 3.53 & \best 3.53 & \second 3.36 \\

\bottomrule
\end{tabular}
}
\end{table}

\FloatBarrier

\subsubsection{Radar-Camera Depth Estimation.}

Radar sensing has gained increasing attention in autonomous driving and robotics for its low cost, all-weather capability, and long effective range. The task of radar–camera depth estimation aims to recover dense metric depth by fusing RGB frames with mmWave Radar. However, Radar point clouds are extremely sparse and noisy—over a thousand times sparser than LiDAR, and often misaligned spatiotemporally with other onboard sensors, leading to a substantial modality gap from the sparse metric prompts used in our pretraining.

To ensure fair evaluation, our pretraining datasets intentionally exclude all Radar data to avoid information leakage. We therefore use this task to rigorously assess the transferability of our pretrained model to an unseen sensor. 
\vspace{2pt}
We conduct two experiments on the nuScenes dataset~\cite{caesar2020nuscenes} using the standard split~\cite{wang2025tacodepth,long2021radar,li2024radarcam-depth,lindepth_radar,lo2023rcdpt_radar,sun2024cafnet_radar,singh2023depth_radar}:
(1) training the teacher model from scratch using Radar signals as prompts and LiDAR maps as ground truth; and
(2) finetuning the pretrained teacher model under the same supervision.
For comparison, we include several  Radar–image fusion baselines, including Lin et al. \cite{lindepth_radar},  RC-PDA \cite{long2021radar},  R4Dyn \cite{gasperini2021r4dyn},  DORN \cite{lo2021depth_dorn} , Singh et al. \cite{singh2023depth_radar}, CaFNet \cite{sun2024cafnet_radar}, Li et al. \cite{li2024sparse},  RadarCam-Depth~\cite{li2024radarcam-depth} and TacoDepth~\cite{wang2025tacodepth}.

As shown in Tab.~\ref{tab:radar-comparison}, finetuning our pretrained model with Radar prompt achieves state-of-the-art performance—nearly doubling the accuracy of its from-scratch counterpart and surpassing all prior fusion methods.
These results demonstrate that our pretraining paradigm scales effectively: training with randomly sampled sparse prompts allows the model to learn versatile metric representations, enabling seamless adaptation to new and unseen sensing modalities such as mmWave Radar. In contrast to previous works that focus on training and testing with single sensors, our experimental results suggest that future research should pay more attention to general pretraining paradigms.

\begin{table*}[t!]
\centering
\caption{Radar-Camera Depth Estimation. Comparisons on the nuScenes \cite{caesar2020nuscenes} dataset (in millimeters)}
\footnotesize
\label{tab:radar-comparison}
  \begin{tabular*}{\textwidth}{@{\extracolsep{\fill}}ll|cc|cc|cc@{}}
\toprule[1.5pt]
\multirow{2}{*}{Type} & \multirow{2}{*}{Method} & \multicolumn{2}{c|}{0-50m} & \multicolumn{2}{c|}{0-70m} & \multicolumn{2}{c}{0-80m} \\
\cmidrule(lr){3-4} \cmidrule(lr){5-6} \cmidrule(lr){7-8}
& & MAE$\downarrow$ & RMSE$\downarrow$ & MAE$\downarrow$ & RMSE$\downarrow$ & MAE$\downarrow$ & RMSE$\downarrow$ \\
\midrule
\multirow{8}{*}{\shortstack{\textit{Indep-}\\ \textit{endent}}} 
& Lin \textit{et al.} \cite{lindepth_radar} & 2034.9 & 4316.5 & 2294.7 & 5338.2 & 2371.0 & 5623.0 \\
& RC-PDA \cite{long2021radar} & 2225.0 & 4156.5 & 3326.1 & 6700.6 & 3713.6 & 7692.8 \\
& R4Dyn \cite{gasperini2021r4dyn} & -- & -- & -- & -- & -- & 6434.0 \\
& DORN \cite{lo2021depth_dorn} & 1926.6 & 4124.8 & 2380.6 & 5252.7 & 2467.7 & 5554.3 \\
& Singh \textit{et al.} \cite{singh2023depth_radar} & 1727.7 & 3746.8 & 2073.2 & 4590.7 & 2179.3 & 4898.7 \\
& CaFNet \cite{sun2024cafnet_radar} & 1674.2 & 3674.5 & 2010.3 & 4493.1 & 2109.8 & 4765.6 \\
& Li \textit{et al.} \cite{li2024sparse} & \underline{1524.5} & \underline{3567.3} & \underline{1822.9} & \underline{4303.6} & \underline{1927.0} & \underline{4609.6} \\
& TacoDepth \cite{wang2025tacodepth} & \textbf{1423.6} & \textbf{3275.8} & \textbf{1712.6} & \textbf{3960.5} & \textbf{1833.4} & \textbf{4150.2} \\
\midrule
\multirow{2}{*}{\textit{Plug-in}} 
& RadarCam-Depth \cite{li2024radarcam-depth} & 1286.1 & 2964.3 & 1587.9 & 3662.5 & 1689.7 & 3948.0 \\
& TacoDepth \cite{wang2025tacodepth} & \textbf{1046.8} & \textbf{2487.5} & \textbf{1347.1} & \textbf{3152.8} & \textbf{1492.4} & \textbf{3324.8} \\
\midrule
\multirow{2}{*}{\textbf{Metric Anything-Pretrain}} 
& From-Scratch & 1335.4 & 2958.8 & 1622.9 & 3788.1 & 2101.7 & 4033.2 \\
& Finetune & \textbf{651.4} & \textbf{2084.4} & \textbf{863.6} & \textbf{2771.6} & \textbf{934.5} & \textbf{3057.5} \\
\bottomrule[1.5pt]
\end{tabular*}
\end{table*} 
\FloatBarrier
 
\subsection{Prompt-Free Downstream Tasks}  

\subsubsection{Monocular Depth Estimation.}

Metric prediction can be broadly divided into two tracks: (1) Monocular metric depth map estimation and (2) Monocular geometry estimation. The first track aims to predict a dense point map where each pixel encodes the metric distance from the camera origin to the visible surface point along its viewing ray. The second track is monocular geometry estimation, which focuses on predicting a per-pixel 3D point map in the camera coordinate system.

\begin{table}[t]
\centering
\tiny
\caption{Monocular depth estimation performance of our Student-DepthMap against prior methods, measured by $\delta_1$ accuracy (\%, $\uparrow$).}
\label{tab:student_mde_vs}
\resizebox{1\columnwidth}{!}{
\begin{tabular}{lccccccc}
\toprule
Method 
& ETH3D 
& Booster 
& NuScenes 
& Sun-RGBD 
& Sintel 
& Middlebury 
& Rank $\downarrow$ \\
\midrule

DepthAnything \cite{depth_anything_v1}    
& 9.3        & \third 52.3  & 35.4       & 85.0       & 6.9        & 39.3       & 6.00 \\

DAV2 \cite{depth_anything_v2} 
& 36.3       & \best 59.5   & 17.7       & 72.4       & 5.9        & 37.2       & 6.17 \\

Metric3D \cite{yin2023metric3d}    
& 34.2       & 4.7        & 64.4       & 16.9       & 17.3       & 13.6       & 6.83 \\

Metric3D-v2 \cite{hu2024metric3d}    
& \best 87.7 & 39.4       & \third 82.6 & 75.6       & \second 38.3 & 29.9       & \third 4.17 \\

PatchFusion \cite{li2024patchfusion}    
& \third 51.8 & 22.6       & 20.4       & 53.6       & 14.0       & 49.9       & 6.00 \\

UniDepth \cite{piccinelli2024unidepth}    
& 25.3       & 27.6       & \second 83.6 & \second 95.8 & 16.5       & 31.9       & 5.00 \\

ZoeDepth \cite{bhat2023zoedepth}    
& 34.2       & 21.6       & 28.1       & 85.7       & 7.8        & \third 53.8  & 5.83 \\

DepthPro \cite{depthpro}    
& 41.5       & 46.6       & 49.1       & \third 89.0 & \best 40.0  & \second 60.5 & \second 3.17 \\
\midrule

\textbf{Student-DepthMap}
& \second 79.9 & \best 59.5 & \best 88.1 & \best 97.7 & \third 27.7 & \best 65.8 & \best 1.50 \\
\bottomrule
\end{tabular}
}
\end{table}

\begin{table*}[h]
\centering
\caption{\textbf{Monocular Depth Estimation on Diverse Unseen Datasets.} Zero-shot evaluation across six benchmarks covering indoor, outdoor, driving, and synthetic domains. Our method consistently  ranks \textbf{1st} or \textbf{2nd}  across all six datasets.
}

\label{tab:more_metrics_on_student_depthmap}
\footnotesize

\begin{minipage}[t]{0.49\textwidth}
\centering
\textbf{ETH3D \cite{schops2017eth3d}}
\vspace{2pt}
\resizebox{\textwidth}{!}{
\begin{tabular}{l|cccc}
\toprule
\textbf{Method} & \textbf{AbsRel}$\downarrow$ & \textbf{Log$_{10}$}$\downarrow$ & $\boldsymbol{\delta_2}$↑ & $\boldsymbol{\delta_3}$↑ \\
\midrule
DepthAnything~\cite{depth_anything_v1} & 1.682 & 0.380 & 19.78 & 31.06 \\
DepthAnything v2~\cite{depth_anything_v2} & 0.370 & 0.173 & 64.66 & 86.26 \\
Metric3D~\cite{yin2023metric3d} & 0.859 & 0.240 & 49.29 & 57.57 \\
Metric3D v2~\cite{hu2024metric3d} & \best 0.124 & \best 0.053 & \best 99.55 & \best 99.90 \\
PatchFusion~\cite{li2024patchfusion} & \third 0.256 & \third 0.106 & \third 88.38 & \third 97.31 \\
UniDepth~\cite{piccinelli2024unidepth} & 0.457 & 0.186 & 57.67 & 81.48 \\
ZoeDepth~\cite{bhat2023zoedepth} & 0.500 & 0.176 & 64.45 & 81.43 \\
Depth Pro~\cite{depthpro} & 0.327 & 0.193 & 61.31 & 71.23 \\
\midrule
\textbf{Ours-Student-Depthmap} & \second 0.147 & \second 0.064 & \second 97.71 & \best 99.90 \\
\bottomrule
\end{tabular}
}
\end{minipage}
\hfill
\begin{minipage}[t]{0.49\textwidth}
\centering
\textbf{nuScenes \cite{caesar2020nuscenes}}
\vspace{2pt}
\resizebox{\textwidth}{!}{
\begin{tabular}{l|cccc}
\toprule
\textbf{Method} & \textbf{AbsRel}$\downarrow$ & \textbf{Log$_{10}$}$\downarrow$ & $\boldsymbol{\delta_2}$↑ & $\boldsymbol{\delta_3}$↑ \\
\midrule
DepthAnything~\cite{depth_anything_v1} & 0.453 & 0.151 & 73.88 & 90.30 \\
DepthAnything v2~\cite{depth_anything_v2} & 0.614 & 0.326 & 31.84 & 47.27 \\
Metric3D~\cite{yin2023metric3d} & 0.422 & 0.132 & 77.22 & 83.61 \\
Metric3D v2~\cite{hu2024metric3d} & \third 0.197 & \third 0.080 & \second 93.25 & \third 95.74 \\
PatchFusion~\cite{li2024patchfusion} & 0.392 & 0.226 & 48.74 & 76.04 \\
UniDepth~\cite{piccinelli2024unidepth} & \best 0.138 & \best 0.060 & \third 93.01 & \second 96.42 \\
ZeroDepth~\cite{guizilini2023zerodepth} & 0.237 & 0.121 & 82.60 & 89.91 \\
ZoeDepth~\cite{bhat2023zoedepth} & 0.498 & 0.182 & 64.95 & 82.70 \\
Depth Pro~\cite{depthpro} & 0.287 & 0.164 & 73.84 & 84.25 \\
\midrule
\textbf{Ours-Student-Depthmap} & \second 0.152 & \second 0.063 & \best 96.56 & \best 98.26 \\
\bottomrule
\end{tabular}
}
\end{minipage}

\par\vspace{10pt}

\begin{minipage}[t]{0.49\textwidth}
\centering
\textbf{Sintel \cite{butler2012sintel}}
\vspace{2pt}
\resizebox{\textwidth}{!}{
\begin{tabular}{l|cccc}
\toprule
\textbf{Method} & \textbf{AbsRel}$\downarrow$ & \textbf{Log$_{10}$}$\downarrow$ & $\boldsymbol{\delta_2}$↑ & $\boldsymbol{\delta_3}$↑ \\
\midrule
DepthAnything~\cite{depth_anything_v1} & 3.973 & 0.559 & 15.42 & 27.28 \\
DepthAnything v2~\cite{depth_anything_v2} & 2.226 & 0.494 & 18.70 & 33.82 \\
Metric3D~\cite{yin2023metric3d} & 1.733 & 0.387 & 32.38 & 44.79 \\
Metric3D v2~\cite{hu2024metric3d} & \best 0.370 & \best 0.216 & \best 62.92 & \best 76.87 \\
PatchFusion~\cite{li2024patchfusion} & \third 0.617 & 0.391 & 35.52 & 51.44 \\
UniDepth~\cite{piccinelli2024unidepth} & 0.869 & 0.301 & 35.72 & 57.26 \\
ZeroDepth~\cite{guizilini2023zerodepth} & 0.703 & 0.491 & 25.63 & 37.08 \\
ZoeDepth~\cite{bhat2023zoedepth} & 0.946 & 0.392 & 22.70 & 44.97 \\
Depth Pro~\cite{depthpro} & \second 0.508 & \third 0.230 & \second 59.25 & \second 71.14 \\
\midrule
\textbf{Ours-Student-Depthmap} & 0.792 & \second 0.227 & \third 50.01 & \third 70.57 \\
\bottomrule
\end{tabular}
}
\end{minipage}
\hfill
\begin{minipage}[t]{0.49\textwidth}
\centering
\textbf{Sun-RGBD \cite{song2015sunrgbd}}
\vspace{2pt}
\resizebox{\textwidth}{!}{
\begin{tabular}{l|cccc}
\toprule
\textbf{Method} & \textbf{AbsRel}$\downarrow$ & \textbf{Log$_{10}$}$\downarrow$ & $\boldsymbol{\delta_2}$↑ & $\boldsymbol{\delta_3}$↑ \\
\midrule
DepthAnything~\cite{depth_anything_v1} & 0.114 & 0.053 & \third 98.81 & \second 99.77 \\
DepthAnything v2~\cite{depth_anything_v2} & 0.182 & 0.070 & 97.65 & 99.46 \\
Metric3D~\cite{yin2023metric3d} & 1.712 & 0.382 & 27.00 & 34.12 \\
Metric3D v2~\cite{hu2024metric3d} & 0.156 & 0.076 & 96.35 & 99.55 \\
PatchFusion~\cite{li2024patchfusion} & 0.466 & 0.961 & 60.15 & 60.65 \\
UniDepth~\cite{piccinelli2024unidepth} & \third 0.087 & \third 0.037 & \best 99.33 & \best 99.80 \\
ZoeDepth~\cite{bhat2023zoedepth} & 0.123 & 0.053 & 97.95 & 99.51 \\
Depth Pro~\cite{depthpro} & \second 0.113 & \second 0.049 & 98.51 & 99.55 \\
\midrule
\textbf{Ours-Student-Depthmap} & \best 0.085 & \best 0.033 & \second 99.31 & \third 99.65 \\
\bottomrule
\end{tabular}
}
\end{minipage}

\par\vspace{10pt}

\begin{minipage}[t]{0.49\textwidth}
\centering
\textbf{Middlebury \cite{scharstein2014middlebury}}
\vspace{2pt}
\resizebox{\textwidth}{!}{
\begin{tabular}{l|cccc}
\toprule
\textbf{Method} & \textbf{AbsRel}$\downarrow$ & \textbf{Log$_{10}$}$\downarrow$ & $\boldsymbol{\delta_2}$↑ & $\boldsymbol{\delta_3}$↑ \\
\midrule
DepthAnything~\cite{depth_anything_v1} & 0.273 & 0.149 & 69.62 & 86.06 \\
DepthAnything v2~\cite{depth_anything_v2} & 0.262 & 0.141 & 72.07 & 90.55 \\
Metric3D~\cite{yin2023metric3d} & 1.251 & 0.305 & 37.53 & 58.73 \\
Metric3D v2~\cite{hu2024metric3d} & 0.450 & 0.152 & 73.32 & 88.61 \\
PatchFusion~\cite{li2024patchfusion} & \third 0.250 & \third 0.108 & \third 87.17 & \third 98.15 \\
UniDepth~\cite{piccinelli2024unidepth} & 0.324 & 0.127 & 80.05 & \second 99.62 \\
ZeroDepth~\cite{guizilini2023zerodepth} & 0.377 & 0.179 & 67.06 & 78.95 \\
ZoeDepth~\cite{bhat2023zoedepth} & 0.214 & 0.115 & 77.68 & 90.86 \\
Depth Pro~\cite{depthpro} & 0.251 & \second 0.089 & \second 93.17 & 96.40 \\
\midrule
\textbf{Ours-Student-Depthmap} & \best 0.200 & \best 0.082 & \best 96.17 & \best 99.93 \\
\bottomrule
\end{tabular}
}
\end{minipage}
\hfill
\begin{minipage}[t]{0.49\textwidth}
\centering
\textbf{Booster \cite{ramirez2023booster}}
\vspace{2pt}
\resizebox{\textwidth}{!}{
\begin{tabular}{l|cccc}
\toprule
\textbf{Method} & \textbf{AbsRel}$\downarrow$ & \textbf{Log$_{10}$}$\downarrow$ & $\boldsymbol{\delta_2}$↑ & $\boldsymbol{\delta_3}$↑ \\
\midrule
DepthAnything~\cite{depth_anything_v1} & 0.317 & 0.114 & \second 79.62 & \third 95.23 \\
DepthAnything v2~\cite{depth_anything_v2} & \third 0.315 & \third 0.110 & 76.24 & 94.28 \\
Metric3D~\cite{yin2023metric3d} & 1.332 & 0.346 & 13.07 & 33.98 \\
Metric3D v2~\cite{hu2024metric3d} & 0.417 & 0.140 & 75.78 & 92.83 \\
PatchFusion~\cite{li2024patchfusion} & 0.719 & 0.213 & 49.39 & 72.89 \\
UniDepth~\cite{piccinelli2024unidepth} & 0.500 & 0.166 & 60.90 & 89.21 \\
ZoeDepth~\cite{bhat2023zoedepth} & 0.610 & 0.195 & 52.66 & 75.51 \\
Depth Pro~\cite{depthpro} & 0.336 & 0.118 & \third 79.43 & \second 96.52 \\
\midrule
\textbf{Ours-Student-Depthmap} & \best 0.282 & \best 0.100 & \best 84.11 & \best 96.83 \\
\bottomrule
\end{tabular}
}
\end{minipage}

\end{table*}

\paragraph{Monocular metric depth map.}

\begin{figure}
    \centering
    \includegraphics[width=0.95\linewidth]{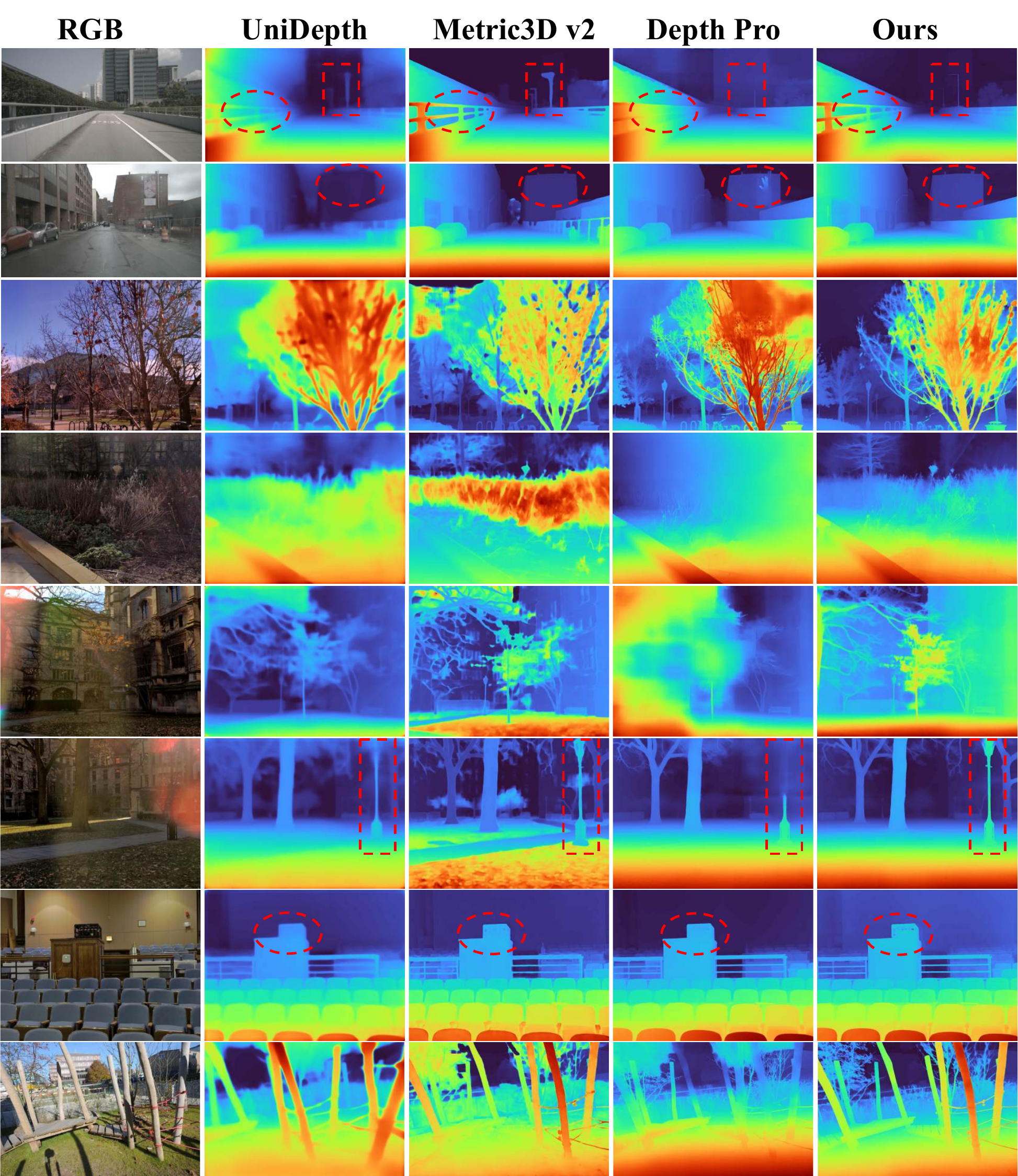}
    \caption{\textbf{Zero-shot Visual Comparisons on Challenging Test Samples.} Our model robustly captures details of thin structures and in scenes with difficult lighting where competitors often fail.}
\label{fig:student_vs_anno}
\end{figure}
\begin{figure}[!t]
    \centering
    \includegraphics[width=0.95\linewidth]{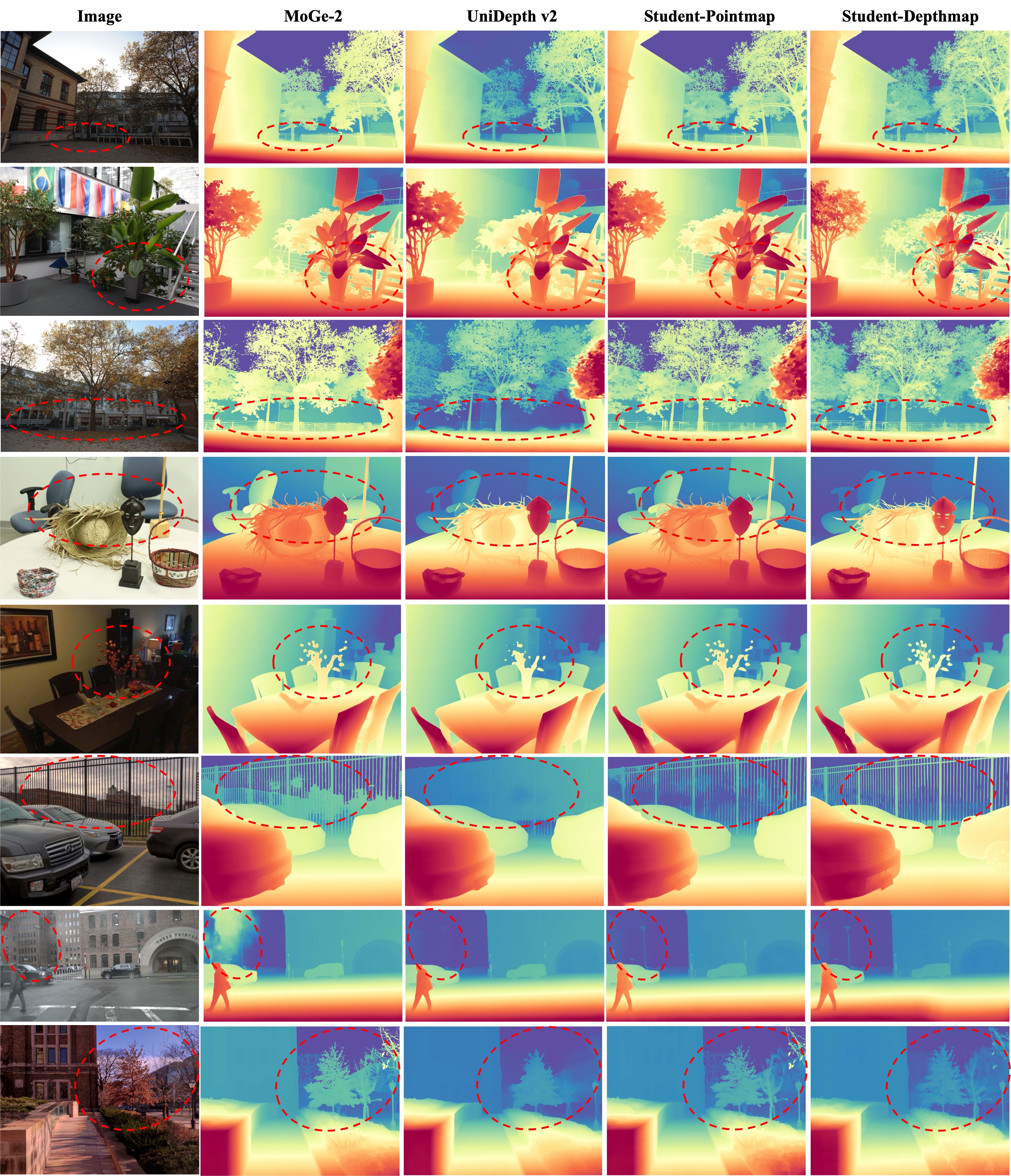}
    \caption{\textbf{Qualitative Comparison of Monocular Depth Estimation}. Compared with MoGe2 and UniDepthv2, our distilled model produces more detailed and geometrically plausible predictions for both depth maps and point maps.}
\label{fig:student-depthmap-vs}
\end{figure}

For metric depth map prediction, we design a dedicated student network (Sec.\ref{sec:student}) and train it from scratch, denoted as `Student-DepthMap' in Tab.~\ref{tab:student_mde_vs} and Tab.~\ref{tab:more_metrics_on_student_depthmap}.
For comparison, we evaluate our method against a wide range of recent approaches, including Depth Anything~\cite{depth_anything_v1}, Depth Anything V2~\cite{depth_anything_v2}, Metric3D~\cite{yin2023metric3d}, Metric3D V2~\cite{hu2024metric3d}, PatchFusion~\cite{li2024patchfusion}, UniDepth~\cite{piccinelli2024unidepth}, ZoeDepth~\cite{bhat2023zoedepth}, and DepthPro~\cite{depthpro}.
We adopt the $\delta_1$ metric, which measures the proportion of inlier pixels based on relative error thresholds, and report results on six datasets — Booster~\cite{ramirez2023booster}, Middlebury~\cite{scharstein2014middlebury}, Sun-RGBD~\cite{song2015sunrgbd}, ETH3D~\cite{schops2017eth3d}, NuScenes~\cite{caesar2020nuscenes}, and Sintel~\cite{butler2012sintel}.
These datasets cover indoor/outdoor scenes, cartoon or game scenes, street views and extreme situations such as severe weather and lighting conditions.
Notably, none of them are used for training, validation, or hyper-parameter tuning — all results are evaluated under a strict zero-shot protocol. Comprehensive evaluation metrics, including AbsRel~\cite{ladicky2014pulling_absrel}, Log$_{10}$ error, $\delta_2$/$\delta_3$ thresholds, and RMSE are provided in the Tab.~\ref{tab:more_metrics_on_student_depthmap}. Quantitative results are presented Fig.~\ref{fig:student_vs_anno} and Fig.~\ref{fig:student-depthmap-vs}.


To ensure a strict zero-shot protocol, we exclude any model–dataset pair if the evaluation set was used during model training. Performance is measured using standard depth estimation metrics: AbsRel, $\log_{10}$, and threshold accuracy scores ($\delta_1$, $\delta_2$, $\delta_3$). 
As reported in Tab.~\ref{tab:student_mde_vs} and Tab.~\ref{tab:more_metrics_on_student_depthmap}, our method exhibits remarkable robustness and consistency across diverse domains:

\begin{itemize}
    \item \textbf{Indoor \& Complex Geometry:} On \textbf{Sun-RGBD} and \textbf{Middlebury}, which feature cluttered indoor scenes and high-resolution details respectively, our method achieves the best performance (Rank 1), significantly outperforming Metric3D v2 and Depth Pro (e.g., AbsRel \textbf{0.085} vs.\ 0.156 on Sun-RGBD). This indicates our model's superior capability in resolving intricate local geometry in constrained spaces.
    
    \item \textbf{Outdoor \& Driving:} On \textbf{nuScenes} and \textbf{ETH3D}, our method remains highly competitive, ranking 2nd in absolute error metrics while often achieving the highest accuracy in stricter thresholds ($\delta_3$). However, since these GT datasets inherently contain noise, we consider $\delta_3$ metric to be more reasonable and robust. This suggests that our model effectively handles large-scale depth variations typical of outdoor environments.
    
    \item \textbf{Robustness on Unconventional Data:} On \textbf{Booster}, a dataset known for challenging lighting and textures, our method outperforms all baselines (AbsRel \textbf{0.282}), highlighting its resilience to domain shifts. Even on the synthetic \textbf{Sintel} dataset, where domain gaps are significant, we maintain strong performance (2nd best in Log$_{10}$), demonstrating that our learned representations generalize well beyond photorealistic domains.
\end{itemize}

Overall, while some baselines excel in specific niches, our method delivers the most balanced and consistently high performance across the full spectrum of test scenarios. 

\begin{figure}[!t]
    \centering
    \includegraphics[width=0.95\linewidth]{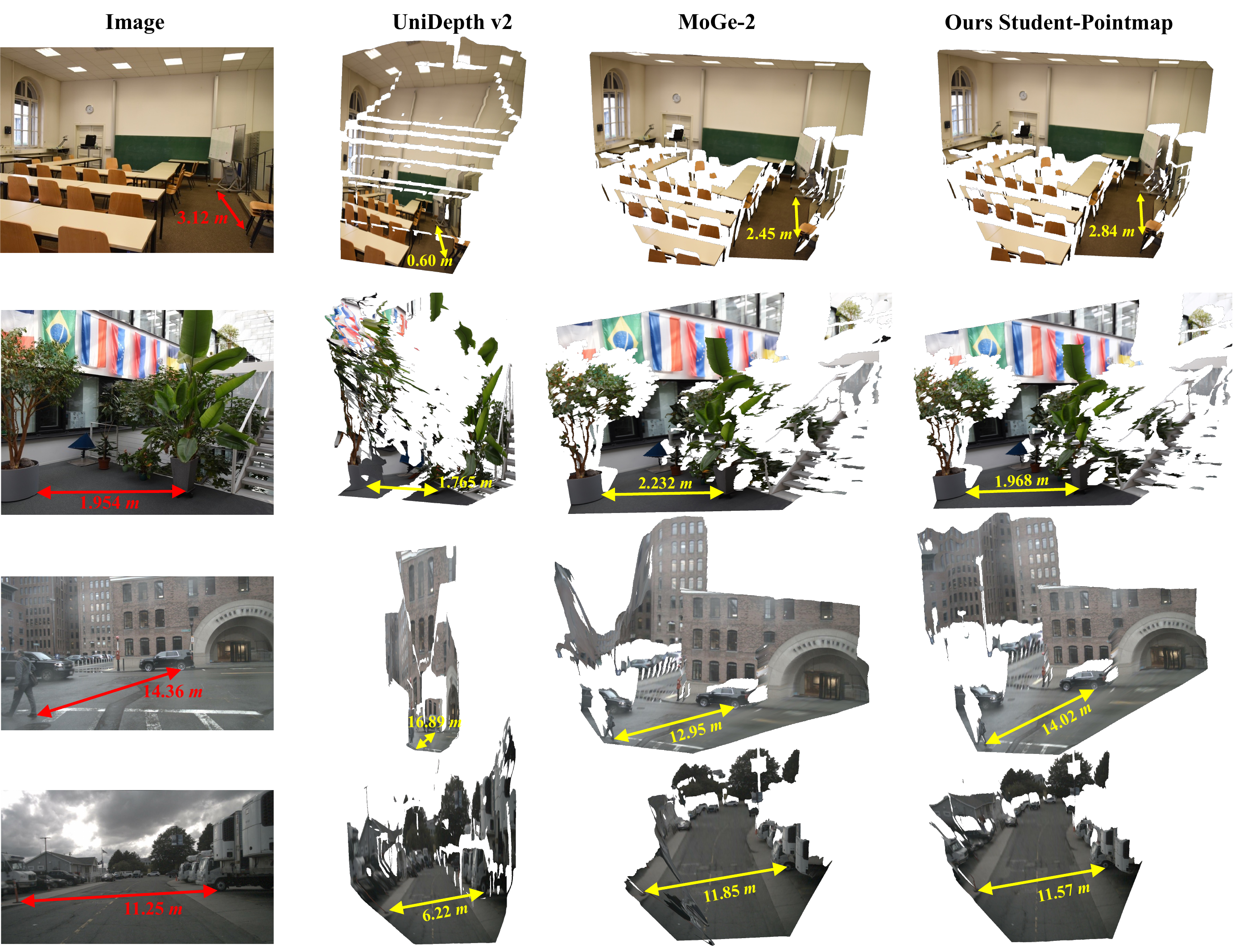}
    \caption{\textbf{Qualitative Comparison of Point Maps.} The \nocolorbox{red}{red arrows} indicate the GT  distance, the \nocolorbox{yellow}{yellow arrows} indicate the distance from predicted point map.}
\label{fig:metric-3d-pointmap-vs}
\end{figure}

\paragraph{Monocular metric point map.}

For metric point map prediction, we leverage pseudo-labels generated by our pre-trained model to fine-tune recent state-of-the-art methods such as MoGe-2~\cite{wang2025moge2}, denoted as `Student-PointMap'.
This design allows us to assess the generality and precision of our pseudo-labels under different training paradigms — whether training from scratch or fine-tuning, and regardless of whether the output head predicts depth maps or 3D point maps.

\begin{figure}[!b]
    \centering
    \includegraphics[width=1\linewidth]{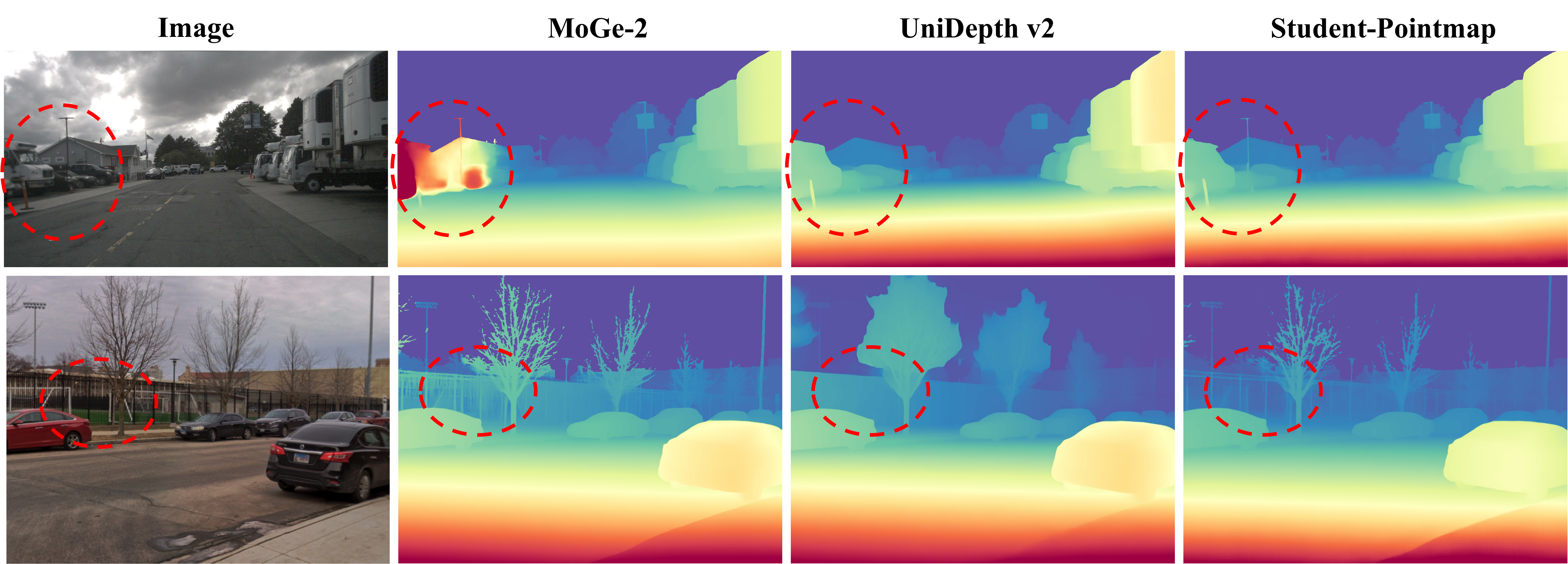}
    \caption{\textbf{Qualitative Comparison of  Depth Map}. Compared with MoGe2 and UniDepthv2, our Student-PointMap model,  which is finetuned from MoGe2 using pseudo-labels predicted by the our  pretrained model, achieves more stable and accurate depth estimation.}
\label{fig:student-pointmap-vs}
\end{figure}

Monocular 3D geometry estimation aims to recover a per-pixel 3D point map in the camera coordinate system. In this setting, we leverage pseudo-labels generated by our pre-trained model to fine-tune recent state-of-the-art frameworks such as MoGe-2~\cite{wang2025moge2}, denoted as “Student-PointMap”. This design enables a comprehensive evaluation of the generality and precision of our pseudo-labels across different training paradigms—including training from scratch versus fine-tuning, and varying output representations (depth maps or 3D point maps). As shown in Fig.~\ref{fig:metric-3d-pointmap-vs}, Fig.~\ref{fig:student-pointmap-vs}, Tab.~\ref{tab:student-pointmap-vs-wo-in} and  Tab.~\ref{tab:student-pointmap-vs-w-in},  our distillation approach consistently achieves state-of-the-art performance, demonstrating its robustness to differences in prediction heads and network initialization. We adopt the GIANT-LARGE and DA3MONO-LARGE variants from the official DepthAnything3~\cite{depthanything3} checkpoints, which represent the largest and most powerful models that support monocular metric depth estimation. However, the performances of GIANT-LARGE and DA3MONO-LARGE are not particularly satisfactory in our setting.  We conjecture that the capability of DepthAnything3~\cite{depthanything3} still depends on inferring matching relationships across multiple views, making accurate metric scale recovery particularly challenging in complex monocular settings.  Additionally, Depth Anything3~\cite{depthanything3} does not support inference when the camera's intrinsic parameters are unknown, which limits its applicability. In contrast, we evaluated its performance under both known and unknown intrinsic parameter conditions.

\begin{table*}[h]
\centering
\caption{\textbf{Monocular Geometry Estimation without Camera Intrinsics.} Our \textit{Student-PointMap} model achieves the best average rank (1.88), demonstrating superior robustness under unknown camera parameters.}

\label{tab:student-pointmap-vs-wo-in}
\resizebox{\textwidth}{!}{
\begin{tabular}{l|cc|cc|cc|cc|ccc}
\hline
\multirow{2}{*}{Method} & \multicolumn{2}{c|}{KITTI \cite{geiger2012_kitti}} & \multicolumn{2}{c|}{ETH3D \cite{schops2017eth3d}} & \multicolumn{2}{c|}{iBims-1 \cite{koch2018ibims-1}} & \multicolumn{2}{c|}{DIODE \cite{vasiljevic2019diode}} & \multicolumn{3}{c}{Avg.} \\
& AbsRel$\downarrow$ & $\delta_1\uparrow$ & AbsRel$\downarrow$ & $\delta_1\uparrow$ & AbsRel$\downarrow$ & $\delta_1\uparrow$ & AbsRel$\downarrow$ & $\delta_1\uparrow$ & AbsRel$\downarrow$ & $\delta_1\uparrow$ & Rank$\downarrow$ \\
\hline
ZoeDepth \cite{bhat2023zoedepth} & 17.0 & 85.4 & 57.1 & 33.7 & 17.4 & 67.2 & 39.3 & 29.3 & 32.70 & 53.90 & 7.00 \\
MASt3R \cite{leroy2024mast3r} & 56.7 & 9.84 & 47.2 & 20.1 & 18.7 & 61.5 & 54.9 & 19.0 & 44.38 & 27.61 & 9.13 \\
DAV1 \cite{depth_anything_v1} & 11.6 & 94.5 & 40.2 & 24.0 & 12.9 & 81.8 & 58.0 & 16.2 & 30.68 & 54.13 & 6.88 \\
DAV2 \cite{depth_anything_v2} & 10.6 & 88.6 & 36.1 & 36.3 & \second 11.1 & \second 91.7 & 41.2 & 22.1 & 24.75 & 59.68 & 4.88 \\
UniDepth V1 \cite{piccinelli2024unidepth} & \second 4.69 & \best 98.4 & 56.9 & 14.9 & 23.8 & 57.6 & \second 17.1 & \second 71.9 & 25.62 & 60.70 & 5.50 \\
UniDepth V2 \cite{piccinelli2025unidepthv2} & \third 8.58 & \third 95.4 & 20.7 & 69.5 & \best 9.52 & \best 93.2 & 43.0 & 51.8 & 20.45 & \second 77.48 & \second 3.63 \\
Depth Pro \cite{depthpro} & 23.5 & 38.3 & 38.5 & 32.8 & 15.9 & 81.5 & 31.9 & 37.7 & 27.45 & 47.58 & 6.75 \\
MoGe-2 \cite{wang2025moge2} & 18.1 & 62.9 & \best 10.4 & \best 90.8 & 13.6 & 83.0 & \third 17.5 & \third 66.4 & \second 14.90 & \third 75.78 & \third 4.13 \\
DAV3-Metric-Large \cite{depthanything3} & 11.8 & 83.0 & \third 11.0 & \third 90.1 & 27.6 & 43.5 & 24.2 & 53.7 & \third 18.66 & 67.59 & 5.88 \\
DAV3-Nested-G-L \cite{depthanything3}  & 83.2 & 0.00 & 80.7 & 0.00 & 39.1 & 20.9 & 51.9 & 26.4 & 63.73 & 11.84 & 10.38 \\
\midrule
\textbf{Ours-Student-Pointmap} & \best 3.22 & \second 97.7 & \second 10.8 & \second 90.2 & \third 11.3 & \third 86.6 & \best 13.9 & \best 79.8 & \best 9.81 & \best 88.58 & \best 1.88 \\
\hline
\end{tabular}
}
\end{table*}

\begin{table*}[h]
\centering
\caption{\textbf{Monocular Geometry Estimation with Provided Camera Intrinsics.} When ground-truth intrinsics are provided, our \textit{Student-PointMap} model significantly outperforms the baseline metric depth estimation models.}

\label{tab:student-pointmap-vs-w-in}
\resizebox{\textwidth}{!}{
\begin{tabular}{l|cc|cc|cc|cc|ccc}
\hline
\multirow{2}{*}{Method} & \multicolumn{2}{c|}{KITTI \cite{geiger2012_kitti}} & \multicolumn{2}{c|}{ETH3D \cite{schops2017eth3d}} & \multicolumn{2}{c|}{iBims-1 \cite{koch2018ibims-1}} & \multicolumn{2}{c|}{DIODE \cite{vasiljevic2019diode}} & \multicolumn{3}{c}{Avg.} \\
& AbsRel$\downarrow$ & $\delta_1\uparrow$ & AbsRel$\downarrow$ & $\delta_1\uparrow$ & AbsRel$\downarrow$ & $\delta_1\uparrow$ & Rel$\downarrow$ & $\delta_1\uparrow$ & AbsRel$\downarrow$ & $\delta_1\uparrow$ & Rank$\downarrow$ \\
\hline
Metric3D V2 \cite{hu2024metric3d} & \second 5.25 & \third 98.0 & \third 11.8 & \third 88.8 & 9.96 & \second 94.1 & 49.1 & 1.98 & 19.03 & 70.72 & 3.38 \\
UniDepth V1 \cite{piccinelli2024unidepth} & \best 4.43 & \second 98.5 & 44.5 & 26.7 & 22.6 & 60.5 & \third 21.0 & 63.5 & 23.13 & 62.30 & 3.75 \\
UniDepth V2 \cite{piccinelli2025unidepthv2} & 5.98 & 97.7 & 15.0 & 85.2 & \best 7.71 & \best 95.5 & 41.0 & \third 67.1 & \third 17.42 & \third 86.38 & \third 3.13 \\
MoGe-2 \cite{wang2025moge2}  & 8.64 & 93.7 & \second 10.5 & \second 92.2 & \third 9.92 & \third 92.4 & \second 16.2 & \best 77.1 & \second 11.32 & \second 88.85 & \second 2.88 \\
\midrule
\textbf{Ours-Student-Pointmap} & \third 5.47 & \best 99.6 & \best 9.75 & \best 92.4 & \second 9.47 & 92.0 & \best 14.1 & \second 74.3 & \best 9.69 & \best 89.56 & \best 1.88 \\
\hline
\end{tabular}
}
\end{table*}

\subsubsection{Recovering Camera Intrinsics.} Furthermore, we utilize the point map $X$ predicted by our finetuned model (`Student-PointMap' ) to infer the intrinsic parameters of the camera from a straightforward optimization. Throughout our experiments, we assume a unit aspect ratio and that the principal point is approximately centered in the image; therefore, the only unknown intrinsic parameter is the focal length of the first camera, denoted $f$. We estimate $f$ by minimizing a weighted reprojection error,
\begin{equation}
f^{*} = \underset{f}{\arg\min} \sum_{i=0}^{W} \sum_{j=0}^{H}
\left| \left(i^{\prime}, j^{\prime}\right) - f \frac{\left(X_{i,j,0}, X_{i,j,1}\right)}{X_{i,j,2}} \right|,
\label{eq:focal1}
\end{equation}

\noindent where $i'$ and $j'$ are the pixel coordinates expressed relative to the image center, and $X_{i,j,k}$ denotes the $k$-th component of the point map at location $(i,j)$. The optimization in~\eqref{eq:focal1} is efficiently solved with a few iterations of a Weiszfeld-type algorithm \cite{weiszfeld1937point}, yielding the estimated focal length $f^{*}$. We compare our model to optimizated methods using point map \cite{,piccinelli2024unidepth,wang2024dust3r,yin2021learning} and learning-based camera  calibration \cite{jin2023perspective,zhu2023tame}. As shown in Tab.~\ref{tab:fov-comparison}, we achieve the best average accuracy in terms of mean error and median error in degrees. 

 \begin{table*}[t]
\centering
\caption{ Evaluation Results for Camera  Calibration in Degrees.}
\label{tab:fov-comparison}
\begin{tabular*}{\textwidth}{@{\extracolsep{\fill}}l|ccccccc@{}}
\toprule
\multirow{2}{*}{Method} & \multicolumn{2}{c|}{ETH3D \cite{schops2017eth3d}} & \multicolumn{2}{c|}{iBims-1 \cite{koch2018ibims-1}} & \multicolumn{3}{c}{Average} \\
& Mean$\downarrow$ & Med.$\downarrow$ & Mean$\downarrow$ & Med.$\downarrow$ & Mean$\downarrow$ & Med.$\downarrow$ & Rank$\downarrow$ \\
\midrule
Perspective \cite{jin2023perspective} & 13.6 & 11.9 & 10.6 & 9.30 & 12.1 & 10.6 & 7.25 \\
WildCam \cite{zhu2023tame} & 7.70 & 5.81 & 9.48 & 9.08 & 8.59 & 7.45 & 5.00 \\
Depth Pro \cite{depthpro} & 7.18 & 6.34 & 4.12 & 2.58 & 5.65 & 4.46 & 3.50 \\
\midrule
LeReS \cite{yin2021learning} & 8.26 & 7.19 & 18.4 & 17.5 & 13.33 & 12.35 & 7.00 \\
DUSt3R \cite{wang2024dust3r} & 5.77 & 3.60 & 3.83 & \textbf{2.53} & 4.80 & 3.07 & 2.25 \\
UniDepth \cite{piccinelli2024unidepth} & 10.7 & 9.96 & 11.9 & 5.96 & 11.3 & 7.96 & 6.25 \\
MoGe-2 \cite{wang2025moge2} & 4.66 & 3.04 & 8.17 & 6.77 & 6.42 & 4.91 & 3.25 \\
\midrule
\textbf{Student-PointMap} & \textbf{2.50} & \textbf{1.86} & \textbf{3.74} & 3.24 & \textbf{3.12} & \textbf{2.55} & \textbf{1.50} \\
\bottomrule
\end{tabular*}
\end{table*}



\subsubsection{Zero-shot Boundaries Accuracy Measure}
We evaluate the sharpness of the predicted geometry by our `Student-DepthMap' using two synthetic benchmarks, Spring \cite{mehl2023spring} and Sintel \cite{butler2012sintel}, together with the real-world iBims‑1 \cite{koch2018ibims-1} dataset. As reported in Tab.~\ref{tab:boundary_comparison}, our method achieves the best average accuracy boundaris compared to DepthPro \cite{depthpro} and MoGe-2 \cite{wang2025moge2}.

\begin{table}[b!]
\centering
\caption{\textbf{Zero-shot Depth Boundary (F1 score \cite{depthpro}) on Multiple Benchmarks.}}
\label{tab:boundary_comparison}
\begin{tabular*}{\columnwidth}{@{\extracolsep{\fill}}l|cccc@{}}
\toprule
Method & Sintel F1$\uparrow$ & Spring F1$\uparrow$ & iBims F1$\uparrow$ & Rank$\downarrow$ \\
\midrule
\multicolumn{5}{l}{\textit{Relative}} \\
DepthAnything \cite{depth_anything_v1} & 0.261 & 0.0109 & 0.127 & 7.33 \\
DepthAnything v2 \cite{depth_anything_v2} & 0.228 & 0.0610 & 0.111 & 8.00 \\
Marigold \cite{ke2024marigold} & 0.068 & - & 0.149 & 7.50 \\
\midrule
\multicolumn{5}{l}{\textit{Absolute}} \\
DPT \cite{ranftl2021dpt} & 0.181 & - & 0.113 & 9.00 \\
Metric3D \cite{yin2023metric3d} & 0.037 & - & 0.055 & 11.50 \\
Metric3D v2 \cite{hu2024metric3d} & 0.321 & 0.0723 & 0.096 & 6.33 \\
ZoeDepth \cite{bhat2023zoedepth} & 0.027 & 0.0043 & 0.035 & 11.33 \\
PatchFusion \cite{li2024patchfusion} & 0.312 & - & 0.134 & 6.00 \\
UniDepth \cite{piccinelli2024unidepth} & 0.316 & 0.0017 & 0.039 & 8.67 \\
Depth Pro \cite{depthpro} & \best{0.409} & \second{0.1100} & \third{0.176} & \second{2.00} \\
UniDepth-v2 \cite{piccinelli2025unidepthv2} & \third{0.344} & 0.0737 & 0.138 & 4.00 \\
MoGe-v2 \cite{wang2025moge2} & 0.282 & \third{0.0890} & \best{0.194} & \third{3.67} \\
\bottomrule
\textbf{Student-DepthMap} & \second{0.382} & \best{0.1635} & \second{0.179} & \best{1.67} \\

\bottomrule
\end{tabular*}
\end{table}

\begin{figure}
    \centering
    \includegraphics[width=1\linewidth]{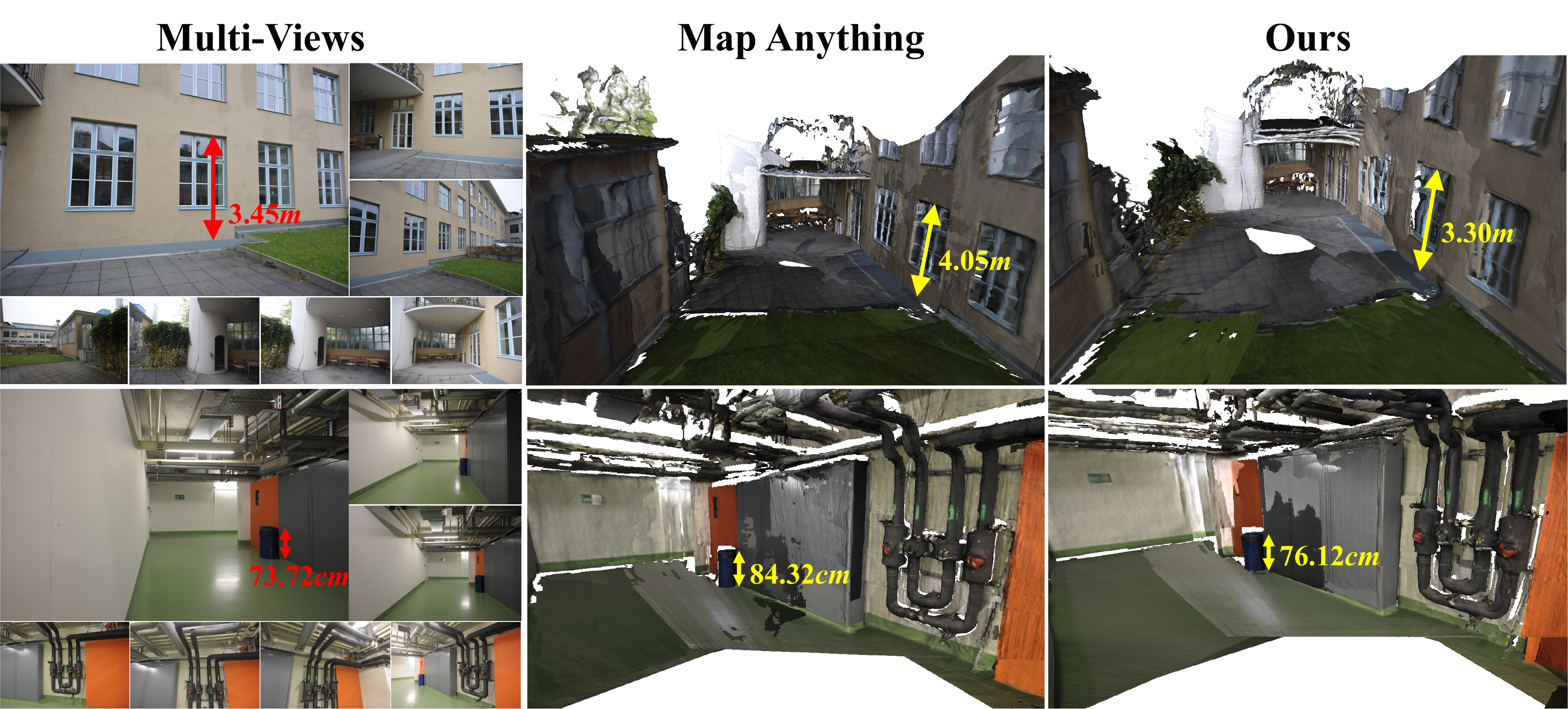}
    \caption{Auxiliary monocular depth inputs improve performance of MapAnything. The \nocolorbox{red}{red arrows} indicate the GT  distance, the \nocolorbox{yellow}{yellow arrows} indicate the distance from 3D reconstruction.}.
\label{fig:mv_recon_vs_mapa}
\end{figure}

\subsubsection{Multi-view Metric 3D Reconstruction.} To validate the accuracy and cross-view consistency of our monocular depth predictions without prompt, we compare with Map Anything \cite{keetha2025mapanything}, a recent state-of-the-art method for multi-view metric 3D reconstruction method.  We use the officially released checkpoint without any additional post-processing or finetuning.  Our evaluation takes the multi-view images together with the per-frame monocular depth maps predicted by our prompt-free model as input. No cross-frame correction  or post-processing optimization techniques (e.g., Bundle Adjustment) is applied, the 3D metric reconstruction results were obtained through a single feed-forward. We report results on the ETH3D \cite{schops2017eth3d} and ScanNet \cite{dai2017scannet} test sets, as show in Tab.~\ref{tab:improve_mapanything}. Compared with the baseline, our approach achieves superior performance, particularly in metric-scale estimation, as show in Fig.~\ref{fig:mv_recon_vs_mapa}. The results show that our proposed pretraining paradigm can transfer its capabilities to a prompt-free student model via distillation. In a prompt-free setting, our model exhibits strong cross-view-consistent depth prediction without relying on any additional conditions. It delivers virtually cost-free improvements in multi-view 3D metric reconstruction accuracy, relying solely on powerful monocular priors rather than multi-view matching-based inference.

\begin{figure}[!b]
    \centering
    \includegraphics[width=0.9\linewidth]{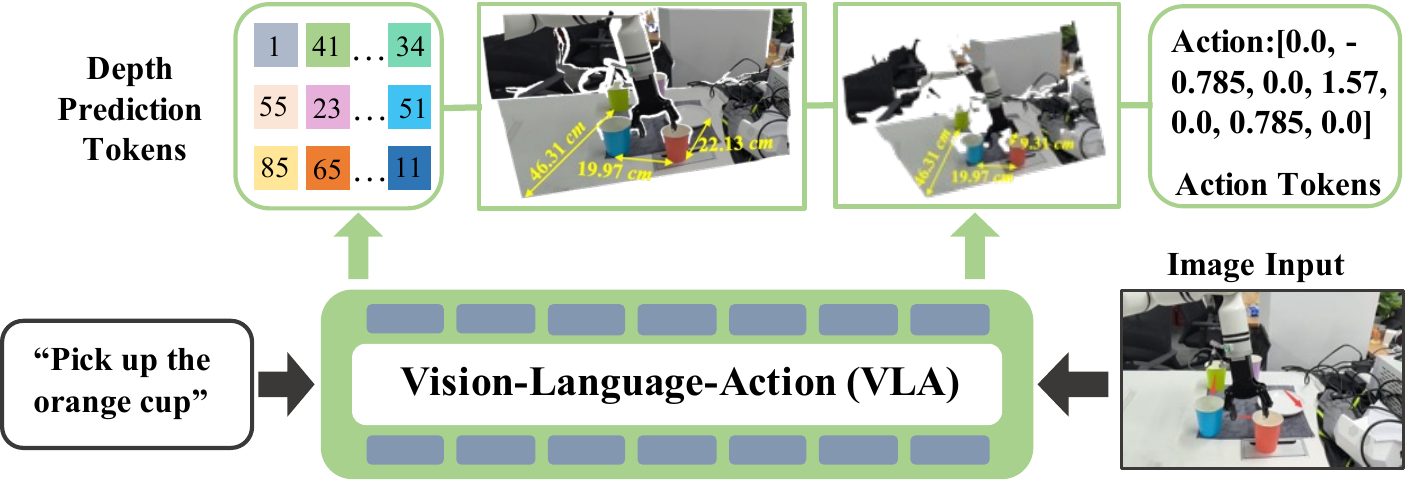}
    \caption{
    \textbf{Enhancing VLA Planning with Metric Anything.} We distill the depth-perception capability of  Metric Anything into the VLA model by supervising it to predict metric-aware depth tokens.}  
\label{fig:vla_app}
\end{figure}

\begin{table*}[t]
\centering
\caption{Integrating our Metric Anything Student module into the MapAnything baseline yields significant performance gains on multi-view metric 3D reconstruction (AbsRel, $\delta_1$ in \%).}
\label{tab:improve_mapanything}
\begin{tabular*}{\textwidth}{@{\extracolsep{\fill}}lcccccc@{}}
\toprule
 \multirow{2}{*}{Method} & \multicolumn{2}{c}{ETH3D \cite{schops2017eth3d}} & \multicolumn{2}{c}{Scannet \cite{dai2017scannet}} & \multicolumn{2}{c}{Avg} \\
 \cmidrule(lr){2-3} \cmidrule(lr){4-5} \cmidrule(lr){6-7}
 & AbsRel$\downarrow$ & $\delta_1\uparrow$ & AbsRel$\downarrow$ & $\delta_1\uparrow$ & AbsRel$\downarrow$ & $\delta_1\uparrow$ \\
 \midrule
MapAnything \cite{keetha2025mapanything}  & 20.43 & 69.07 & 37.61 & 61.78 & 29.02 & 65.42 \\
MapAnything + Ours & \textbf{18.98} & \textbf{73.94} & \textbf{5.90} & \textbf{99.41} & \textbf{12.44} & \textbf{86.68} \\
\bottomrule
\end{tabular*}
\end{table*}

\begin{table}[t]
\centering
\caption{LIBERO benchmark success rates by task category.}
\label{tab:VLA-table}
\begin{tabular*}{\columnwidth}{@{\extracolsep{\fill}}l|cccc|c@{}}
\toprule
\textbf{Method} & \textbf{Spatial} & \textbf{Object} & \textbf{Goal} & \textbf{Long} & \textbf{Avg} \\
\midrule
TraceVLA~\cite{zheng2024tracevla} & 84.6 & 85.2 & 75.1 & 54.1 & 74.8 \\
Octo-Base~\cite{team2024octo} & 78.9 & 85.7 & 84.6 & 51.1 & 75.1 \\
OpenVLA~\cite{kim2024openvla} & 84.7 & 88.4 & 79.2 & 53.7 & 76.5 \\
SpatialVLA~\cite{qu2025spatialvla} & 88.2 & 89.9 & 78.6 & 55.5 & 78.1 \\
CoT-VLA~\cite{zhao2025cot} & 87.5 & 91.6 & 87.6 & 69.0 & 83.9 \\
NORA-AC~\cite{hung2025nora} & 85.6 & 89.4 & 80.0 & 63.0 & 79.5 \\
WorldVLA~\cite{cen2025worldvla} & 87.6 & 96.2 & 83.4 & 60.0 & 79.1 \\
$\pi_0$-FAST~\cite{pertsch2025fast} & \textbf{96.4} & \textbf{96.8} & 88.6 & 60.2 & 85.5 \\
ThinkAct~\cite{huang2025thinkact} & 88.3 & 91.4 & 87.1 & 70.9 & 84.4 \\
\midrule
Baseline-DAV2 \cite{depth_anything_v2} & 87.0 & 95.4 & 87.6 & 77.2 & 86.6 \\
\textbf{Ours} & 88.6 & 94.4 & \textbf{88.8} & \textbf{78.8} & \textbf{87.7} \\
\bottomrule
\end{tabular*}

\end{table}

\subsubsection{VLA Planning} 
We distill our Metric Anything's capability into Vision-Language-Action (VLA) models for action planning. Results (Tab.~\ref{tab:VLA-table}) demonstrate SoTA performance.  3D spatial perception is essential for manipulation in the physical world. Prior work \cite{bhat20253d}  attempt to input depth maps from depth sensors or offline depth-estimation models, along with RGB observations, into vision-language-action (VLA) policies model to improve performance. However, such approaches still require additional hardware or depth predictors at test time, which increases deployment complexity on real robots. Following MolmoACT \cite{lee2025molmoact},  we distill the depth-perception capability of our trained prompt-free model into the VLA model by supervising it to predict depth tokens, rather than consuming depth inputs. We adopt Depth Anything V2 \cite{depth_anything_v2} as a baseline and report success rates on four tasks from the LIBERO benchmark \cite{liu2023libero}. As shown in Tab.~\ref{tab:VLA-table}, our model achieves more accurate spatial understanding of both the environment and the target objects than Depth Anything V2, yielding the best average success rate. Moreover, it achieves significant improvements over models without depth perception or input, indicating that distilling depth perception into VLA policies is a promising avenue for enhancing manipulation performance, even without depth inputs during either training or inference. An overview of the framework is provided in  Fig.~\ref{fig:vla_app}

\subsubsection{Spatial Understanding of MLLMs}
We further evaluate how Metric Anything enhances the 3D spatial reasoning ability of foundation VLMs. 
Metric Anything encodes rich metric 3D information, and we use its pretrained ViT encoder as a spatial perception backbone to provide 3D-aware features for a vision-language model. Concretely, we extract 3D feature tokens from the ViT encoder of Metric Anything and fuse them with the 2D visual tokens of the VLM via a cross-attention module, where the 2D tokens serve as queries and the 3D tokens as keys and values. In this way, the visual stream of the VLM is explicitly conditioned on the metric 3D prior learned by Metric Anything. The fused tokens are then passed through two linear layers to obtain the final fused representation, which is used as the new visual token input to the language model. 
Following VLM-3R~\cite{bhat20253d}, we adopt LLaVA-Next-Video-7B as the base VLM and conduct supervised fine-tuning on 200K general question-answer pairs, and 4,225 embodied route planning instances released by VLM-3R~\cite{bhat20253d}. 

\begin{table*}[b]
\centering
\vspace{1mm}
\captionof{table}{\small \textbf{Evaluation Results on VSI-Bench~\cite{Yang2024ThinkingIS}}. For Spatial-MLLM and Qwen2.5VL-series~\cite{Bai2025Qwen25VLTR}, we use 16 frames as input. For other open-source methods and GPT-4o~\cite{hurst2024gpto}, we follow the setting of VSI-Bench to set frame numbers (ranging from 8 to 32 frames). For Gemini-1.5 Pro~\cite{team2024gemini}, it samples video frames at 1 FPS. \textbf{Bold} denote the best-performing open-source models.}
\setlength\tabcolsep{3pt} 
\resizebox{\textwidth}{!}
{
    \begin{tabular}{l|cccc|cccc|cc}
        \toprule
         \multirow{2}{*}{\textbf{Methods}}  & \multicolumn{4}{c}{\textbf{Numerical Question}} & \multicolumn{4}{c|}{\textbf{Multiple-Choice Question}} & \multirow{2}{*}{\textbf{Avg.}} &\multirow{2}{*}{\textbf{Rank}} \\
        \cmidrule(lr){2-5}\cmidrule(lr){6-9}
         & Obj. Cnt. & Abs. Dist. & Obj. Size & Room Size & Rel. Dist. & Rel. Dir. & Route Plan & Appr. Order & &\\
        \midrule
        \multicolumn{1}{l|}{\textcolor{black}{\textit{Proprietary Models}}} & & & & & & & & & &\\
        GPT-4o~\cite{hurst2024gpto}  & 46.2 & 5.3 & 43.8 & 38.2 & 37.0 & 41.3 & 31.5 & 28.5 & 34.0 & 8\\
        Gemini-1.5 Pro~\cite{team2024gemini} & 56.2 & 30.9 & 64.1 & 43.6 & 51.3 & 46.3 & 36.0 & 34.6 & 45.4 & 3\\
        \midrule
        \multicolumn{1}{l|}{\textcolor{black}{\textit{Open-source Models}}} & & & & & & & & & \\
        InternVL2-40B~\cite{chen2024internvl}  & 34.9 & 26.9 & 46.5 & 31.8 & 42.1 & 32.2 & 34.0 & 39.6 & 36.0 & 7\\
        LongVILA-8B~\cite{Xue2024LongVILASL} & 29.1 & 9.1 & 16.7 & 0.0 & 29.6 & 30.7 & 32.5 & 25.5 & 21.6 & 13\\
        VILA-1.5-40B~\cite{Lin2023VILAOP} & 22.4  & 24.8 & 48.7 & 22.7 & 40.5 & 25.7 & 31.5 & 32.9 & 31.2 & 10\\
        LongVA-7B~\cite{Zhang2024LongCT} & 38.0 & 16.6 & 38.9 & 22.2 & 33.1 & 43.3 & 25.4 & 15.7 & 29.2 & 12\\
        LLaVA-OneVision-72B~\cite{li2024llavaov}  & 43.5 & 23.9 & 57.6 & 37.5 & 42.5 & 39.9 & 32.5 & 44.6 & 40.2 & 5\\
        LLaVA-Video-72B~\cite{Zhang2024VideoIT}  & 48.9 & 22.8 & 57.4 & 35.3 & 42.4 & 36.7 & 35.0 & \textbf{48.6} & 40.9 & 4\\
        Qwen2.5VL-3B~\cite{Bai2025Qwen25VLTR}   & 24.3 & 24.7 & 31.7 & 22.6 & 38.3 & 41.6 & 26.3 & 21.2 & 30.6 & 11\\
        Qwen2.5VL-7B~\cite{Bai2025Qwen25VLTR}  & 40.9 & 14.8 & 43.4 & 10.7 & 38.6 & 38.5 & 33.0 & 29.8 & 33.0 & 9\\
        Qwen2.5VL-72B~\cite{Bai2025Qwen25VLTR} & 25.1 & 29.3 & 54.5 & 38.8 & 38.2 & 37.0 & 34.0 & 28.9 & 37.0 & 6\\
        Spatial-MLLM-4B~\cite{Wu2025SpatialMLLMBM}  & 65.3 & 34.8 & 63.1 & 45.1 & 41.3 & 46.2 & 33.5 & 46.3 & 48.4 & 2\\
        \textbf{Ours}  & \textbf{70.0} & \textbf{51.5} & \textbf{67.5} & \textbf{65.0} & \textbf{66.2} & \textbf{76.8} & \textbf{40.2} & 29.0 & \textbf{58.3} & \textbf{1}\\
        \bottomrule
    \end{tabular}
}
\label{tab:vsibench}
\end{table*}

\begin{figure}
    \centering
    \includegraphics[width=0.7\linewidth]{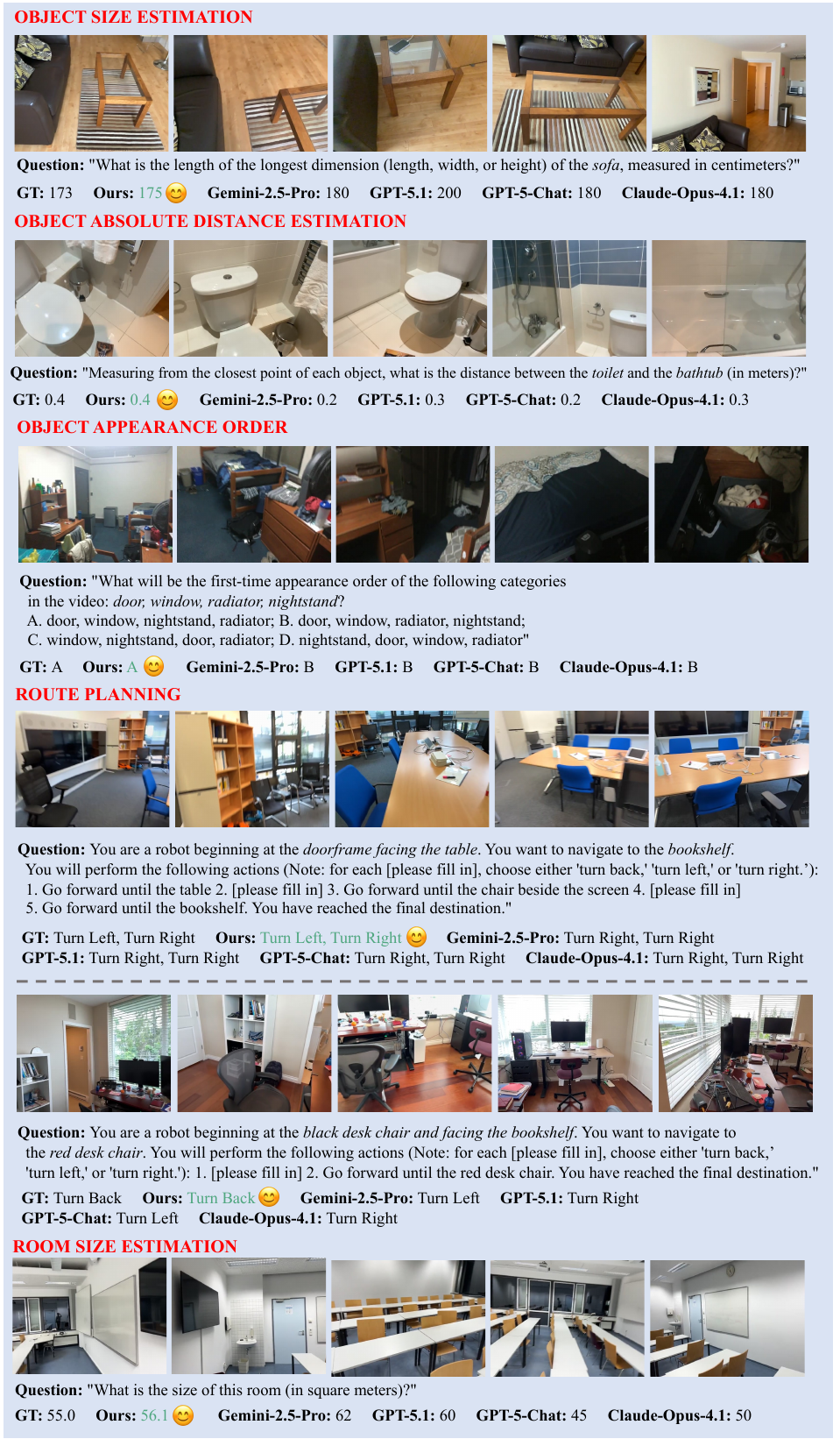}
\caption{\textbf{Enhancing 3D Spatial Reasoning with a Frozen ViT from Metric Anything .}
We evaluate our approach on the VIS Benchmark, covering video question-answering tasks like estimating object size, object's distances, appearance order, route planning, and room size. Compared to mainstream large models, our method demonstrates robust and superior performance in 3D spatial understanding.
}
\label{fig:3d_reasoning}
\end{figure}

\begin{figure}
    \centering
    \includegraphics[width=0.7\linewidth]{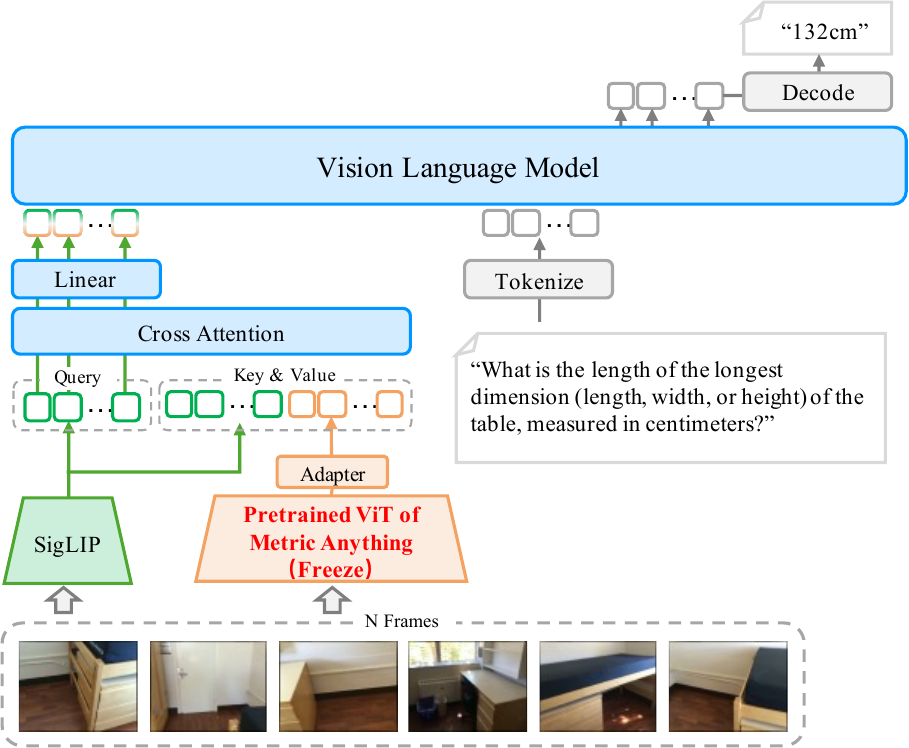}
    \caption{\textbf{Enhancing 3D Spatial Reasoning in MLLMs.} We enhance VLM capabilities by employing the frozen, pretrained ViT from Metric Anything as the visual encoder, thereby preserving its strong spatial understanding during fine-tuning.}
\label{fig:3d_reasoning_framework}
\end{figure}

We report the evaluation results on VSI-Bench~\cite{Yang2024ThinkingIS} in the Tab.\ref{tab:vsibench}. For Spatial-MLLM and Qwen2.5VL-series~\cite{Bai2025Qwen25VLTR}, we use 16 frames as input. For other open-source methods and GPT-4o~\cite{hurst2024gpto}, we follow the setting of VSI-Bench to set frame numbers (ranging from 8 to 32 frames). For Gemini-1.5 Pro~\cite{team2024gemini}, it samples video frames at 1 FPS. As shown in Fig.\ref{fig:3d_reasoning}, we demonstrate various video question-answer categories within the spatial reasoning benchmark (VIS Benchmark), including object size, inter-object distances, object appearance order, route planning, and room size. Quantitative results and case studies demonstrate that the features extracted by our pretrained model encode rich and accurate spatial understanding, and can substantially enhance the spatial reasoning capabilities of existing VLMs. Since spatial understanding is a core competence for tasks such as embodied intelligence, navigation and path planning, and 3D real-world manipulation, we believe that our proposed pretraining paradigm has the potential to serve as a fundamental visual representation backbone that raises the performance ceiling of these tasks. An overview of the framework is provided in  Fig.~\ref{fig:3d_reasoning_framework}. 
\vspace{2mm}

\section{Ablation Study}
\subsection{Scaling up Data.}
We investigate the impact of training data by ablating different proportions of our training dataset. Metric Anything-Pretrain is trained with 5$\%$, 10$\%$, 20$\%$, 40$\%$, 80$\%$, and 100$\%$ of the training set, and its depth super-resolution on $8\times$ downsample performance is evaluated on KITTI, achieving AbsRel values of 5.22, 4.25, 3.98, 3.02, 2.63, 2.34, respectively. 
The results show that while our performance on small-scale datasets lags behind PromptDA, which uses sophisticated rules to simulate low-resolution prompts, our model achieves the best performance as data scales up due to increasing prompt diversity from varied data sources. By minimizing design biases in prompt, our paradigm offers the most robust generalization to downstream tasks without task-specific pre-design.
The data proportion ablation for student is shown in Fig.~\ref{fig:scaleup}. 

\subsection{Network Architecture.} 
Excluding the benefits of data scaling, we evaluate the effectiveness of our ViT-encoder–DPT-head skip-connection design by comparing it with a classic U-Net–style skip-connection architecture \cite{ranftl2021dpt}. For both architectures, we train under two regimes: (1) on noisy, real-world data drawn from multiple sources, and (2) on high-accuracy, domain-aligned pseudo labels produced by our pre-trained model. 
Results in Tab.~\ref{tab:network-comparison} show that classic U-Net skip connections perform well on heterogeneous data but fail to fully leverage ViT's semantic representations when trained with consistent pseudo labels. Our Inverse Skip-Connection design better realizes the potential of pseudo-label distillation. Additionally, even with identical architectures, training with our pseudo labels improves performance, demonstrating our pre-trained model's strong spatial understanding.


\begin{table}[t]
\centering
\caption{Network architecture  ablation on NuScenes and ETH3D.}
\label{tab:network-comparison}
\begin{tabular*}{\linewidth}{@{\extracolsep{\fill}}l|cc|cc|cc|cc@{}}
\toprule
& \multicolumn{4}{c|}{NuScenes \cite{caesar2020nuscenes}} & \multicolumn{4}{c}{ETH3D \cite{schops2017eth3d}} \\
\cmidrule(lr){2-5} \cmidrule(lr){6-9}
Lable Source & \multicolumn{2}{c|}{Real-world} & \multicolumn{2}{c|}{Pseudo} & \multicolumn{2}{c|}{Real-world} & \multicolumn{2}{c}{Pseudo} \\
& Abs$\downarrow$ & RMSE$\downarrow$ & Abs$\downarrow$ & RMSE$\downarrow$ & Abs$\downarrow$ & RMSE$\downarrow$ & Abs$\downarrow$ & RMSE$\downarrow$ \\
\midrule
Unet-sytle & 0.213 & 9.631 & 0.187 & 7.222 & 0.327 & 5.353 & 0.269 & 1.44 \\
\textbf{Ours} & 0.235 & 9.662 & \textbf{0.125} & \textbf{6.267} & 0.334 & 5.490 & \textbf{0.182} & \textbf{1.898} \\
\bottomrule
\end{tabular*}
\end{table}


\subsection{Runtime.} To assess the latency of our pretrained model and prompt-free variants in comparison to baselines, we evaluated all approaches at three image resolutions—VGA (640$\times$480), HD ($1920\times1080$) and 4K ($4032 \times 3024$), and recorded the per-image runtime w/wo prompts. All models were run under FP32 precision in the same environment measured on an H200 GPU. We also report parameter counts and FLOPs for comparison. As summarized in Tab.~\ref{tab:efficiency}, compared to Depth Pro, which shares the most similar in network architecture to ours, our pretrained model does not exhibit a noticeable increase in inference time despite introducing an additional prompt branch, owing to our design that unifies the image encoder and patch encoder into a shared ViT. We further provide two prompt-free variants, including Student-DepthMap and Student-PointMap. Due to their substantially different parameter counts, their inference latencies vary accordingly. Additional results, including boundary evaluation, hyperparameters, training details and loss ablations are provided in the Appendix.

\begin{table}[t!]
\centering
\caption{Single-frame inference latency under FP32 (ms).}
\footnotesize
\label{tab:efficiency}
\begin{tabular*}{\linewidth}{@{\extracolsep{\fill}}lcccccc@{}}
\toprule
\multirow{2}{*}{Method} & \multirow{2}{*}{Para.} & \multirow{2}{*}{Native Res.} & \multirow{2}{*}{FLOPs@HD} & \multicolumn{3}{c}{Latency (ms) @FP32} \\
\cmidrule(lr){5-7}
& &  &  & VGA & HD & 4K \\
\midrule
DepthPro \cite{depthpro} & 952M & $1536\times1536$ & 8848.5G & 246.0 & 246.0 & 246.0 \\
\midrule
Ours-PreTrain & 993M & $1536\times1536$ & 19992.6G & 285.2 & 285.2 & 285.3 \\
Student-PointMap & 326M & $840\times840$ & 4015.7G & 29.4 & 369.8 & 5043.9 \\
Student-DepthMap & 877M & $1536\times1536$ & 21516.4G & 278.8 & 278.5 & 278.8 \\
\bottomrule
\end{tabular*}
\end{table}

\begin{figure}[t]
    \centering
    \includegraphics[width=1\linewidth]{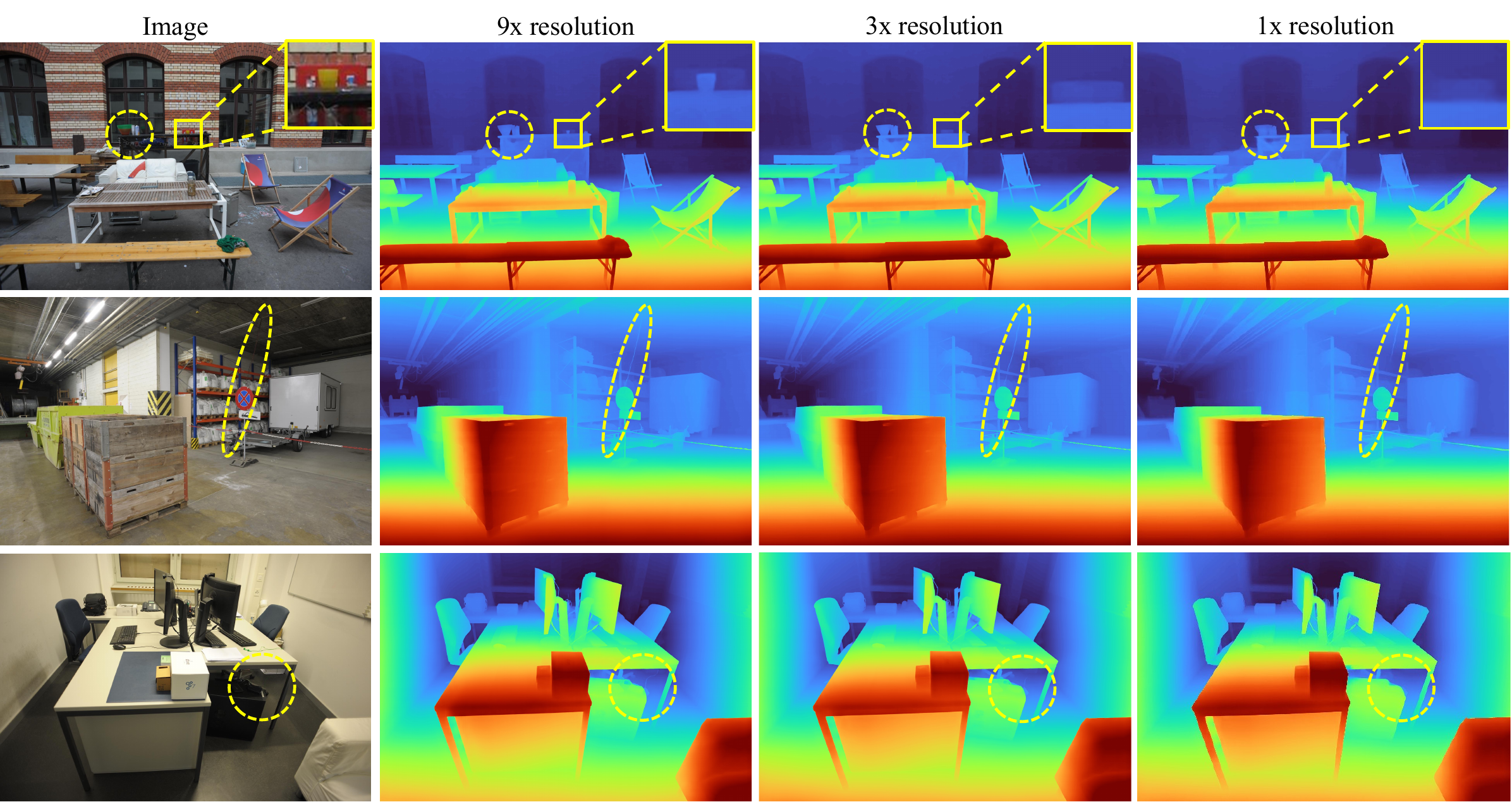}
    \caption{\textbf{Test-time Resolution Scaling}. Qualitative results of  depth estimation on an example image at $1\times$, $3\times$, and $9\times$ the base input resolution. Higher resolutions recover finer details.}
\label{fig:tts}
\end{figure}

\subsection{Test Time Resolution Scaling}
Our metric depth estimation model exhibits a remarkable capability termed ``test-time resolution scaling." This allows the model to process input images at resolutions significantly higher than those encountered during training, which results in progressively refined depth predictions.
As illustrated in Fig.~\ref{fig:tts}, we present depth maps generated at $1\times$, $3\times$, and $9\times$ the base resolution. The results show that increasing the resolution at test time recovers finer structures and high-frequency details, such as thin object boundaries and texture-rich areas. This demonstrates the model's potential for high-resolution depth sensing without the need for fine-tuning.


\subsection{Training Objectives}


To evaluate the impact of different training objectives, we trained our model using losses defined on depth, log-depth, inverse-depth, and our proposed distance-balanced inverse-depth loss  (Eq.~\ref{eq:ours_log_distance} in the main manuscript). Quantitative results in Tab.~\ref{tab:training_objectives} show that our proposed loss achieves performance comparable to the standard inverse-depth loss at close range, while exhibiting superior performance at longer distances. As the depth range increases, the advantage of our method becomes more pronounced. Considering the depth distribution characteristics shown in Fig.~\ref{fig:depthrange} in the main manuscript, and given that our pseudo-labels cover a significantly broader depth range, these results validate the design of our proposed loss function for large-range depth estimation.

\begin{table*}[t]
\centering
\caption{\textbf{Ablation Study for Different Training Objectives under Varying Depth Ranges}. DIODE \cite{vasiljevic2019diode} AbsRel $\downarrow$ metric is reported.}
\label{tab:training_objectives}
\begin{tabular*}{\textwidth}{@{\extracolsep{\fill}}l|cccccc@{}}
\toprule
Training & \multicolumn{6}{c}{DIODE \cite{vasiljevic2019diode} AbsRel $\downarrow$} \\
\cmidrule(lr){2-7}
objective & 0-10m & 10-20m & 10-30m & 30-40m & 40-50m & $>$50m \\
\midrule
Depth \cite{hu2024metric3d}  & 0.574 & 0.582 & 0.599 & 0.622 & 0.645 & 0.689 \\
Log-depth \cite{bhat2023zoedepth} & 0.556 & 0.562 & 0.577 & 0.594 & 0.612 & 0.663 \\
Inverse-depth \cite{depthpro} & 0.467 & 0.493 & 0.565 & 0.581 & 0.592 & 0.632 \\
\textbf{Ours} & \textbf{0.465} & \textbf{0.480} & \textbf{0.489} & \textbf{0.502} & \textbf{0.537} & \textbf{0.589} \\
\bottomrule
\end{tabular*}
\end{table*}

\subsection{Prompt Setting}
We analyze the impact of sparse metric-depth prompt density by varying the number of sampled pixels $N$. Specifically, for each image, we randomly sample 500, 1,000, 2,000, 4,000, 8,000, 16,000, 32,000, or 64,000 valid pixels from the depth map to construct sparse metric prompts. As reported in Tab.\ref{tab:ablations} (left), we present the performance on the Hypersim test set. The results show that as $N$ increases, accuracy gains gradually diminish while computational complexity increases. To balance accuracy and efficiency, we randomly sample \(N \in [2{,}000, 40{,}000]\) valid pixels during training and inference. Notably, although we did not explicitly train for extremely sparse prompts (e.g., N = 100), the benefits of data scaling and the diversity of our collected data enable our pretrained model to maintain state‑of‑the‑art performance, as shown in Tab.\ref{tab:zeroshot_completion} of the main  manuscript.

\begin{table}[htb]
\centering
\small
\caption{\textbf{Two Ablation Studies}: (left) number of prompt points on Hypersim; (right) balance coefficient $C$ on DIODE(AbsRel $\downarrow$).}
\label{tab:ablations}
\begin{tabular*}{\textwidth}{@{\extracolsep{\fill}}ll@{}}
\begin{tabularx}{0.56\textwidth}{l|*{8}{>{\centering\arraybackslash}X}}
\toprule
Points & 500 & 1000 & 2000 & 4000 & 8000 & 16000 & 32000 & 64000 \\
\midrule
AbsRel$\downarrow$ & 0.043 & 0.041 & 0.038 & 0.036 & 0.034 & 0.033 & 0.032 & 0.031 \\
Time (ms) & 224 & 243 & 256 & 266 & 274 & 284 & 299 & 308 \\
\bottomrule
\end{tabularx}
&
\begin{tabularx}{0.4\textwidth}{l|*{5}{>{\centering\arraybackslash}X}}
\toprule
$C$ & 50 & 100 & 200 & 400 & 600 \\
\midrule
0--40\,m & 0.441 & 0.452 & 0.454 & 0.456 & 0.465 \\
$>$40\,m & 0.635 & 0.611 & 0.565 & 0.562 & 0.559 \\
\bottomrule
\end{tabularx}
\end{tabular*}
\end{table}




\subsection{Balance Weights}
\begin{figure*}[t]
    \centering
    \begin{minipage}{0.48\linewidth}
        \centering
        \includegraphics[width=\linewidth]{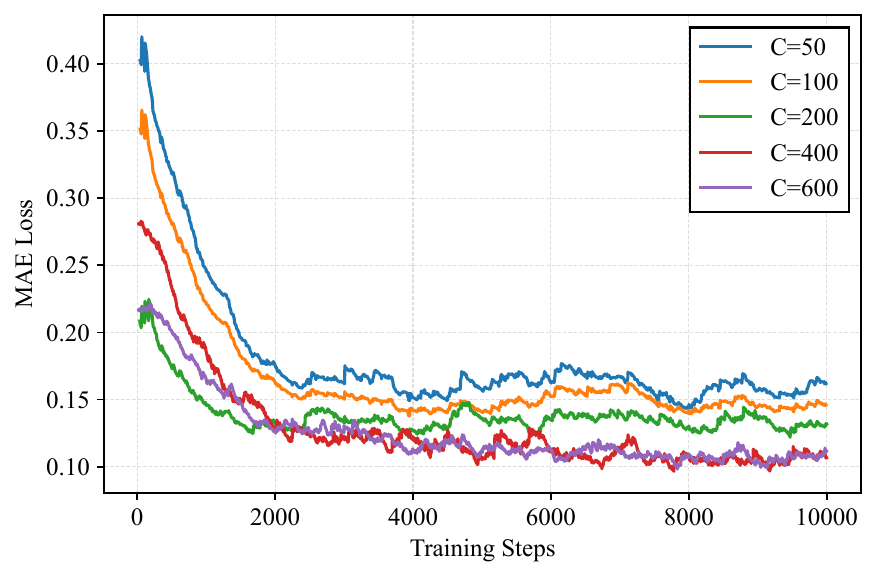}
    \end{minipage}
    \hfill 
    \begin{minipage}{0.5\linewidth}
        \centering
        \includegraphics[width=\linewidth]{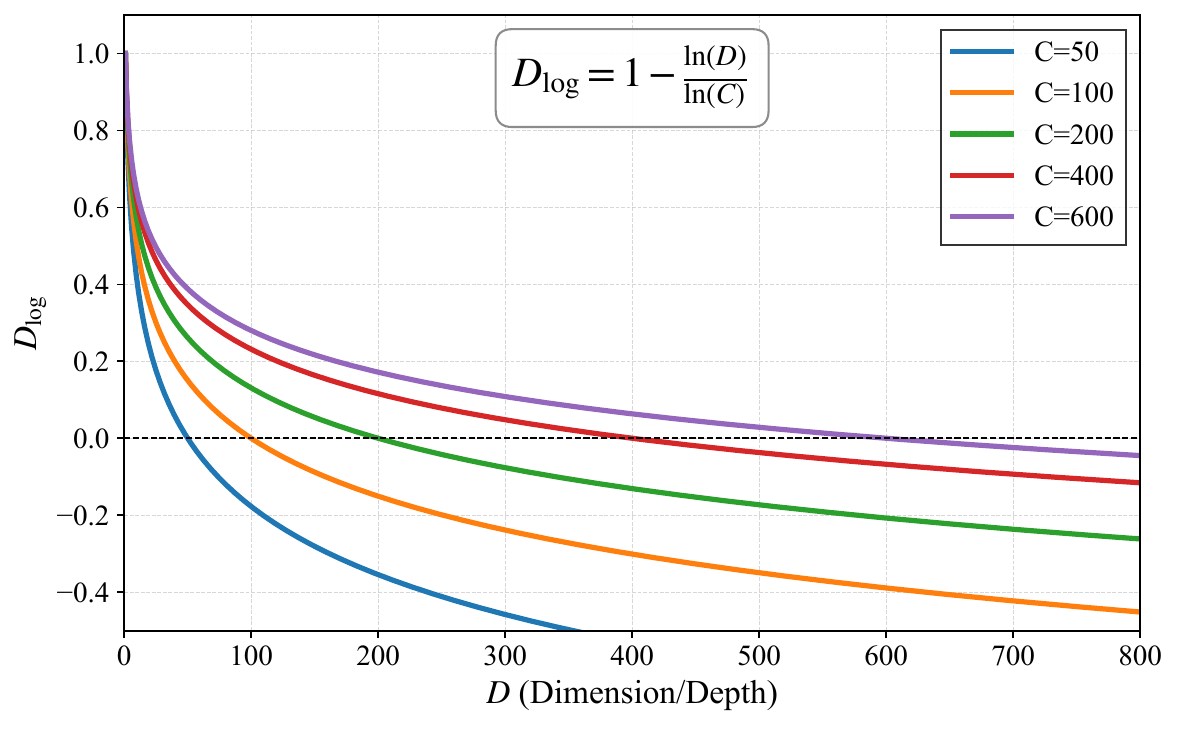}
    \end{minipage}
    \caption{\textbf{Analysis of Our Proposed Distance-Balanced Loss.} Left: The training loss for different $C$ values. Right: The loss function curves. (Eq.~\ref{eq:ours_log_distance} in the main manuscript ).}
    \label{fig:supp_c_loss_analysis}
\end{figure*}

We explore the effect of the balance weight $C$ in Eq.\ref{eq:student_loss} in the main manuscript.  We try different candidates including  $\{50, 100, 200, 400, 600\}$ , and report the results in Tab.\ref{tab:ablations} (right). As shown in Fig.\ref{fig:supp_c_loss_analysis}, if the weight is too large, e.g., 600, it tends to encourage the network to learn depth at distant regions while neglecting details in nearby areas. Conversely, when the weight is too small, the network tends to focus on near-field geometric details at the expense of supervision for distant regions. We therefore set this hyperparameter to $C = 400$, which provides a reasonable trade-off.

\begin{figure}
    \centering
    \includegraphics[width=0.932\linewidth]{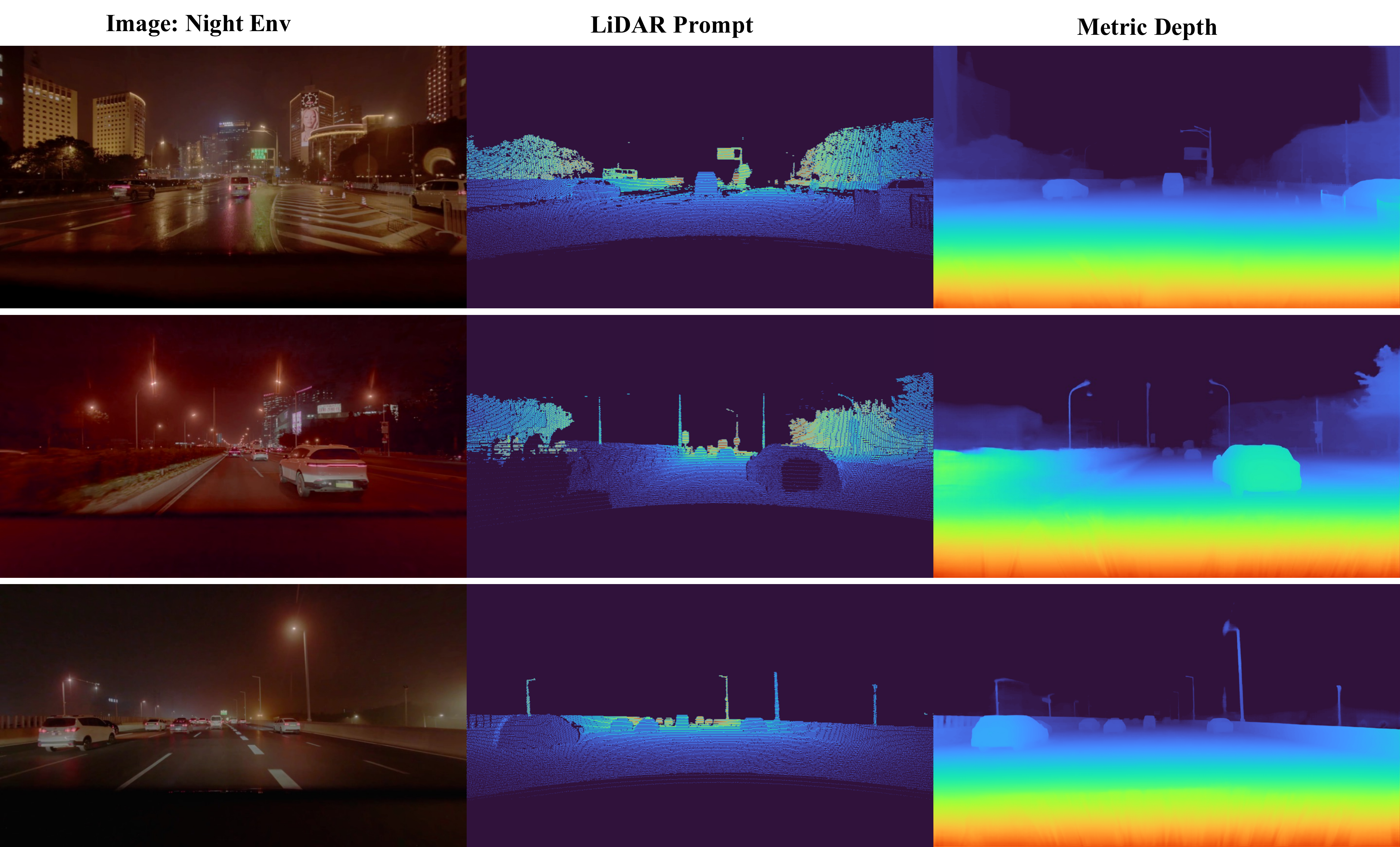}
    \caption{\textbf{Robustness in Night-Time Driving}. We deployed a test vehicle to evaluate performance under low-light conditions. As scene brightness drops, visual signals deteriorate and object details fade. Despite this severe degradation, our model maintains remarkably robust.}
    
\label{fig:supp_ad_night_dc}
\end{figure}

\section{Generalizability to Unseen Sensors, Scenarios, and Extreme Environmental Conditions}

\subsection{Generalization across Sensor Configurations}
\begin{figure}
    \centering
    \includegraphics[width=0.9\linewidth]{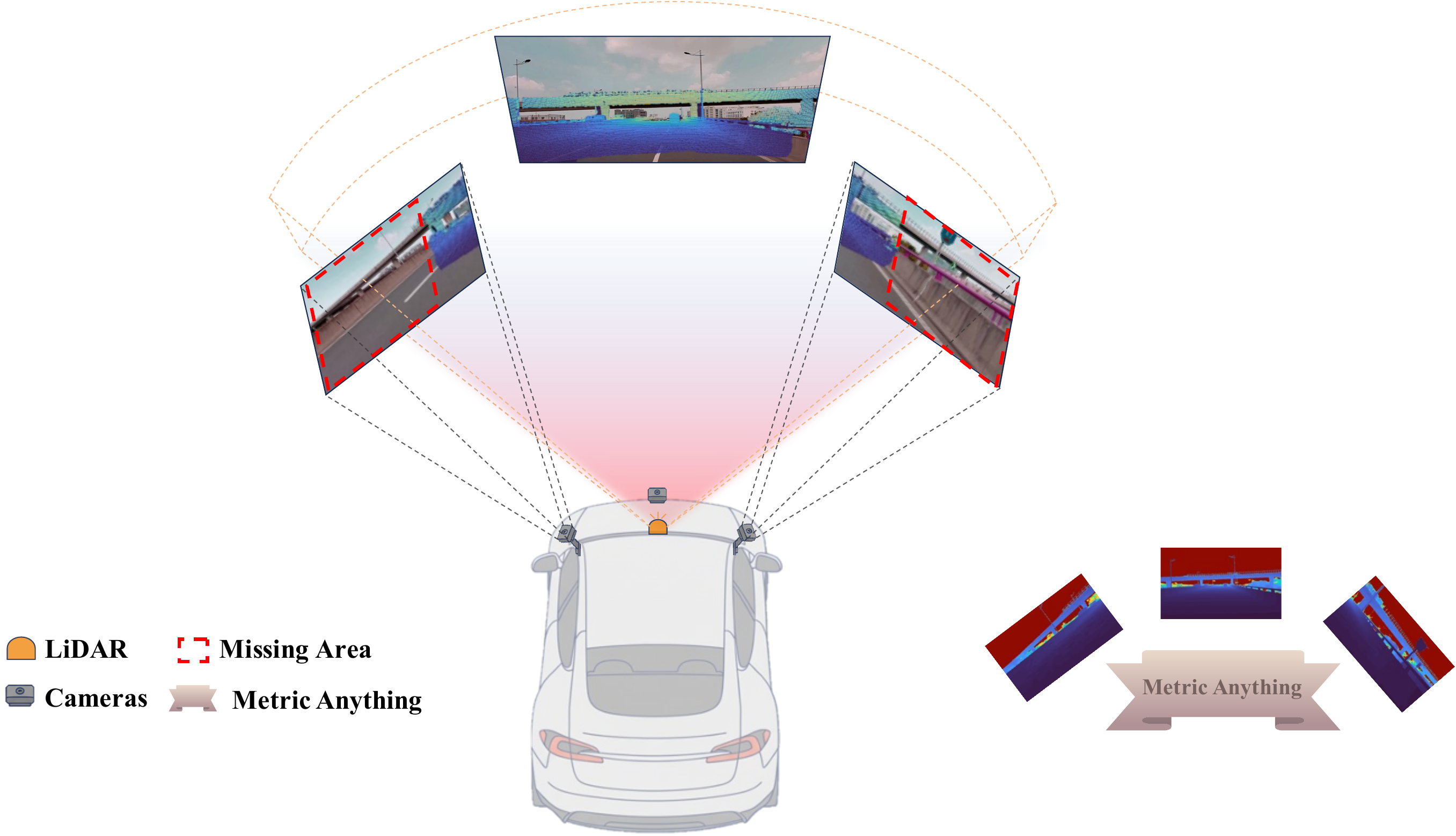}
    \caption{\textbf{Sensor Configuration for Real-World Generalization Evaluation.} 
Our real-world  test vehicle is equipped with three cameras (front, left-front, right-front) and a 128-beam solid-state LiDAR. Due to the LiDAR’s limited vertical field of view(pitch angle limitation), its captured point cloud does not fully cover the cameras’ combined frustums, leaving large image regions without metric depth cues.}
    \label{fig:ad_3c1l}
\end{figure}

\begin{figure}
    \centering
    \includegraphics[width=\linewidth]{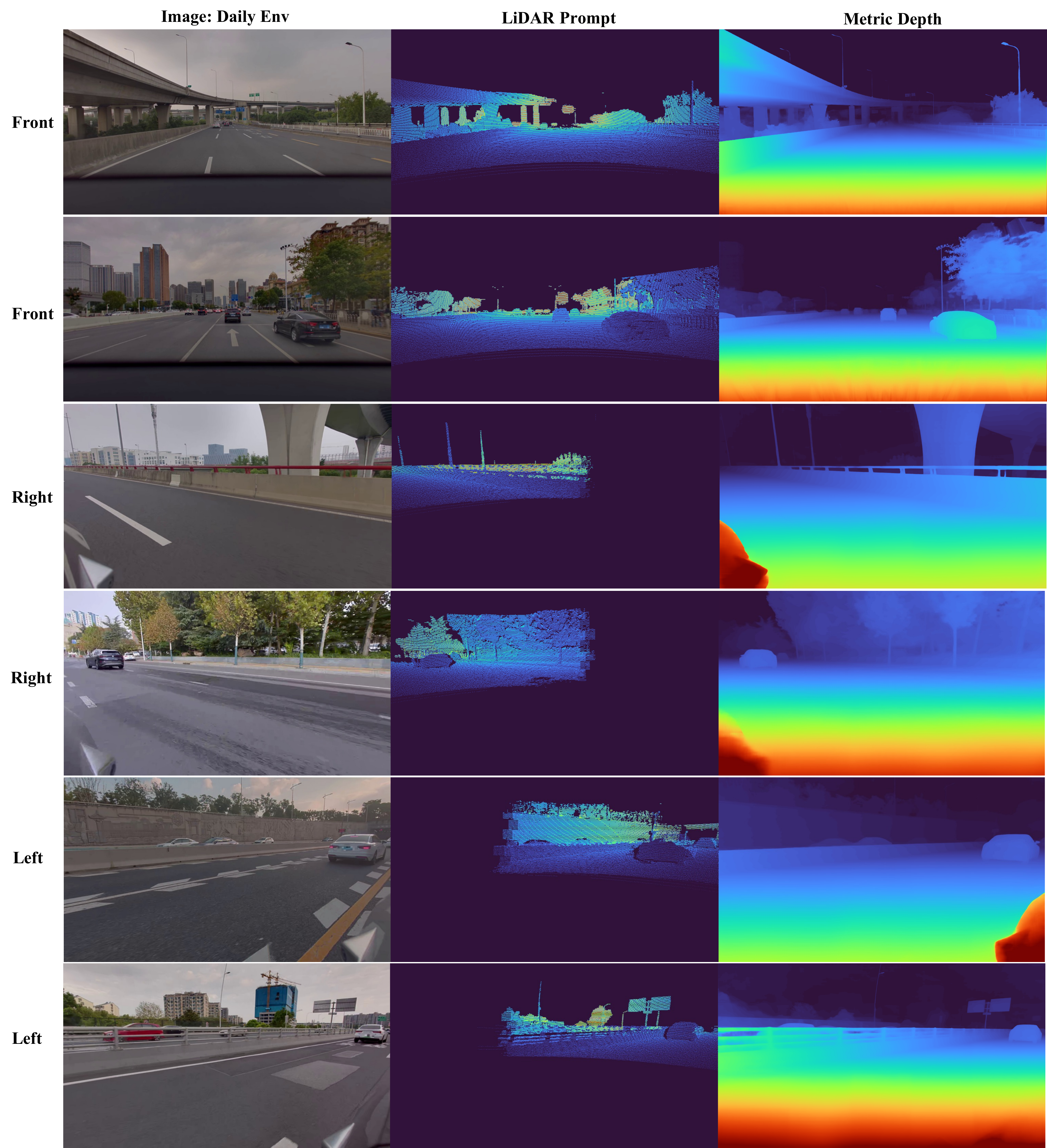}
    \caption{\textbf{Generalization to Real-World Sensor Configurations}. We deployed a test vehicle to evaluate in-the-wild depth super-resolution and completion performance of our pre-trained model without any fine-tuning.}
\label{fig:supp_ad_daily_dc}
\end{figure}

    

This subsection assesses the model's generalization capability to variations in sensor hardware configuration and data characteristics. We deployed a test vehicle equipped with a sensor suite that differed from the training set in both type and spatial arrangement. The setup consisted of three cameras providing front, right-front, and left-front views, coupled with a 128-beam solid-state LiDAR for forward scene perception (see Fig.~\ref{fig:ad_3c1l}).
The collected real-world data exhibits two key challenges: 1) minor calibration inaccuracies and asynchronous sampling rates—with cameras operating at 24 Hz and LiDAR at 10 Hz—introduced spatiotemporal misalignments between sensor modalities; 2) the LiDAR's field of view did not fully cover the lateral areas captured by the side-facing cameras. We deliberately avoided additional post-processing techniques, such as motion compensation, to rigorously evaluate the model's inherent robustness under these realistic imperfections.
The model's performance on two critical tasks is visualized in Fig.~\ref{fig:supp_ad_daily_dc}: \textbf{depth completion} for the lateral blind spots (left-front and right-front views) and \textbf{super-resolution} for the front view. Together, these results demonstrate that our model can faithfully recover the scene's metric depth even when presented with imperfect, real-world data from an unseen sensor configuration.

\subsection{Robustness under Environmental Degradation}

This subsection examines the model's robustness under conditions where environmental interference degrades perceptual signals. 
Two typical scenarios of signal degradation were considered:
\begin{itemize}
\item \textbf{Night-time driving:} Night-time environments introduce multiple challenges including significantly reduced signal-to-noise ratios, loss of texture and color information, over-saturation from artificial light sources, and high-contrast shadows. These factors substantially impact the reliability of vision-based perception systems.
\item \textbf{Rainy/Foggy weather conditions:} LiDAR sensors suffer from reflectivity issues that produce anomalous signals or artifacts. This scenario tests whether our model can rely on visual signals to generate reasonable predictions when LiDAR inputs are corrupted.
\end{itemize}

\begin{figure}
    \centering
    \includegraphics[width=\linewidth]{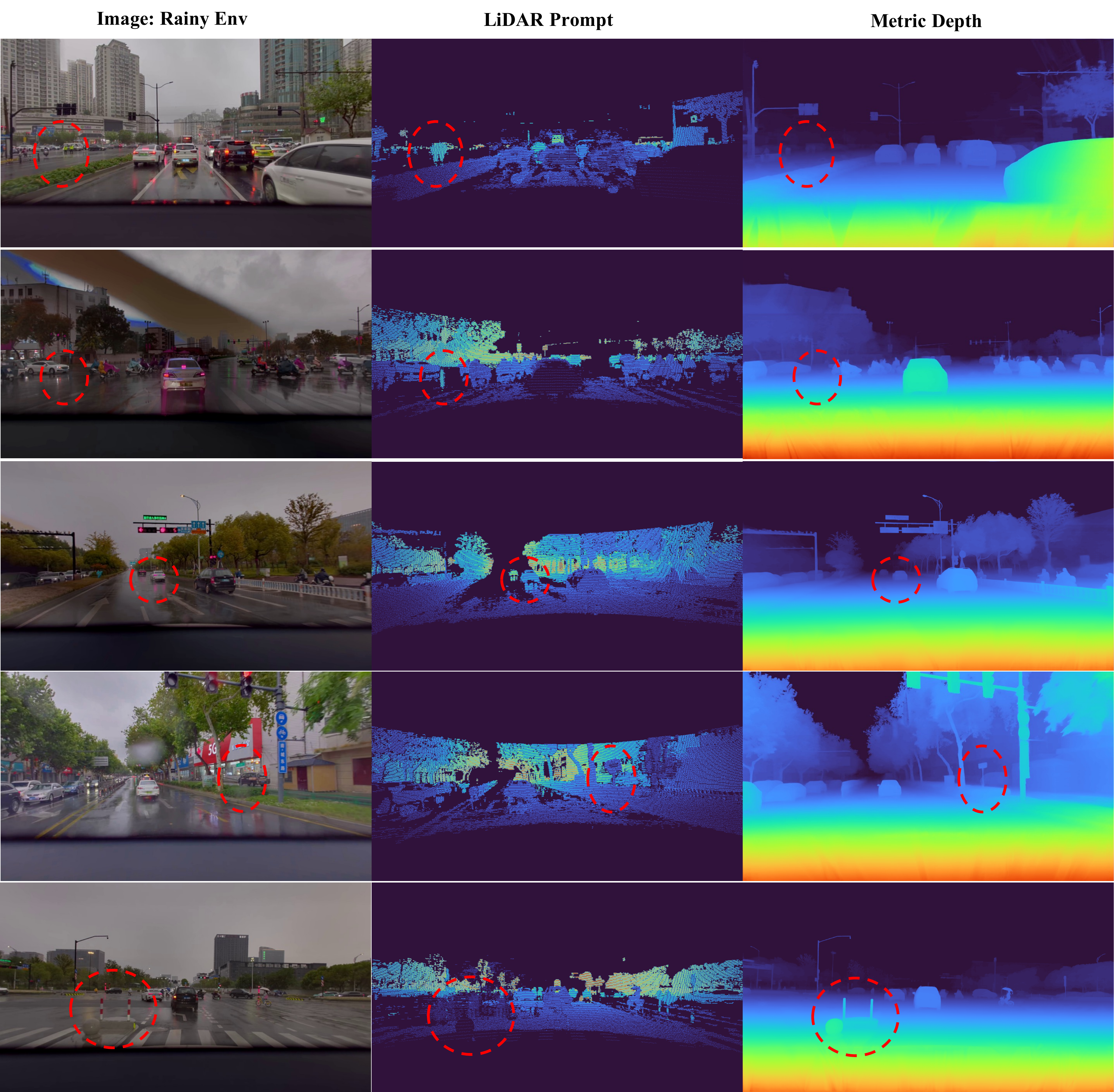}
\caption{\textbf{Robustness in Adverse Weather}. In the real-world deployment, we used a test vehicle to evaluate our pre-trained model for depth super-resolution and completion in rainy and foggy weather conditions without fine-tuning. These adverse conditions significantly affect scene reflectance, causing the LiDAR to produce numerous artifacts or completely occlude critical objects. For example, the degraded data can lead to flat ground surfaces being misinterpreted as uneven or crucial obstacles like pillars being missed. However, our model robustly ignores these erroneous inputs and generates accurate depth predictions based on visual cues, thereby demonstrating the complementary strengths of the two sensing modalities.}
\label{fig:supp_ad_rainy_env_dc}
\end{figure}

As shown in Fig.~\ref{fig:supp_ad_rainy_env_dc} and Fig.~\ref{fig:supp_ad_night_dc}, our model maintains reliable depth estimation in both scenarios, demonstrating strong robustness against environmental degradation. \textbf{ The supplementary video further shows the stability of long-term temporal predictions in our real-world application. }

\subsection{Generalization to Unseen Visual Domains}

\begin{figure}[h]
    \centering
    \includegraphics[width=0.95\linewidth]{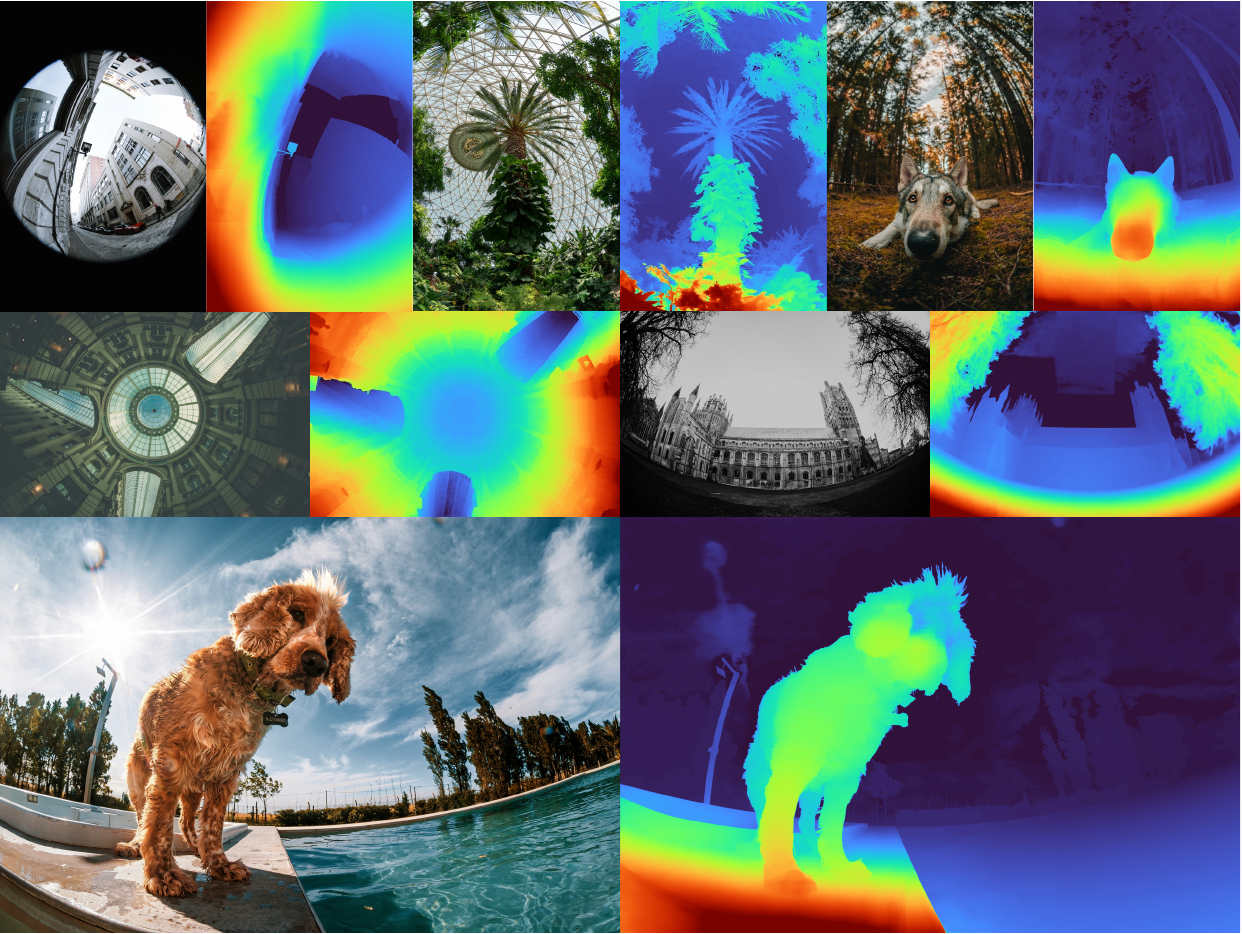}
\caption{\textbf{Generalization to Unseen Visual Domains.} Depth prediction results on \textbf{fisheye images}, an unseen domain characterized by severe radial distortion. The model was applied in a zero-shot setting without fine-tuning.}
\label{fig:supp_vis_fisheye}
\end{figure}

\begin{figure}[h]
    \centering
    \includegraphics[width=0.932\linewidth]{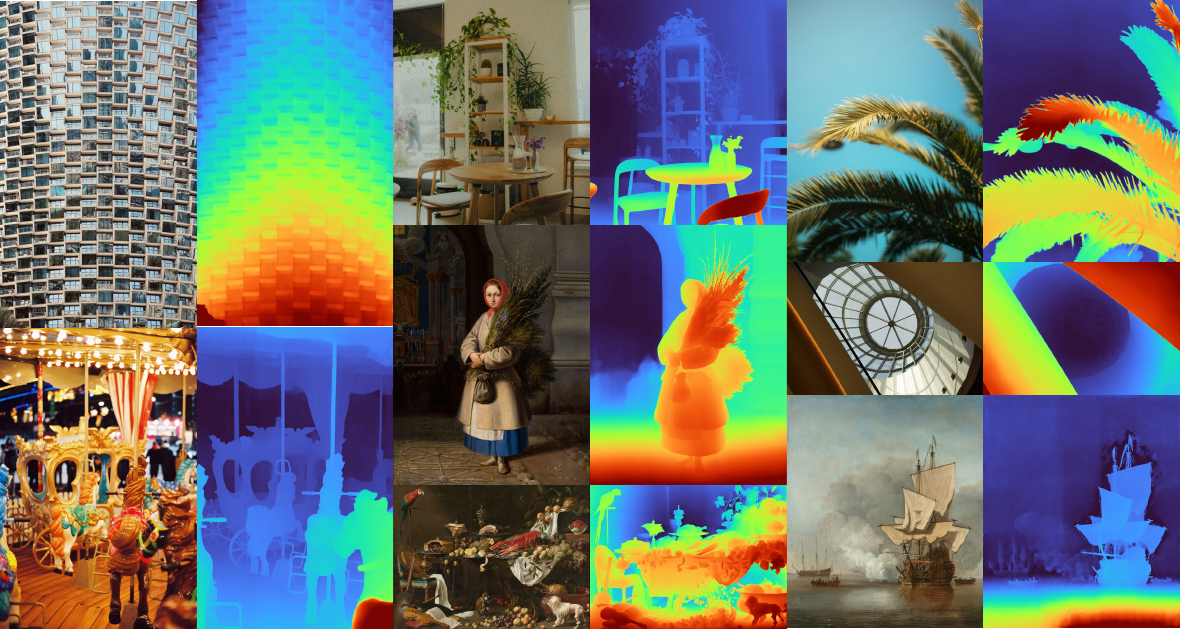}
    \caption{\textbf{Generalization to Unseen Visual Domains.} Depth prediction visualization for diverse in-the-wild images.}
\label{fig:supp_vis_anything2}
\end{figure}

\begin{figure}
    \centering
    \includegraphics[width=0.95\linewidth]{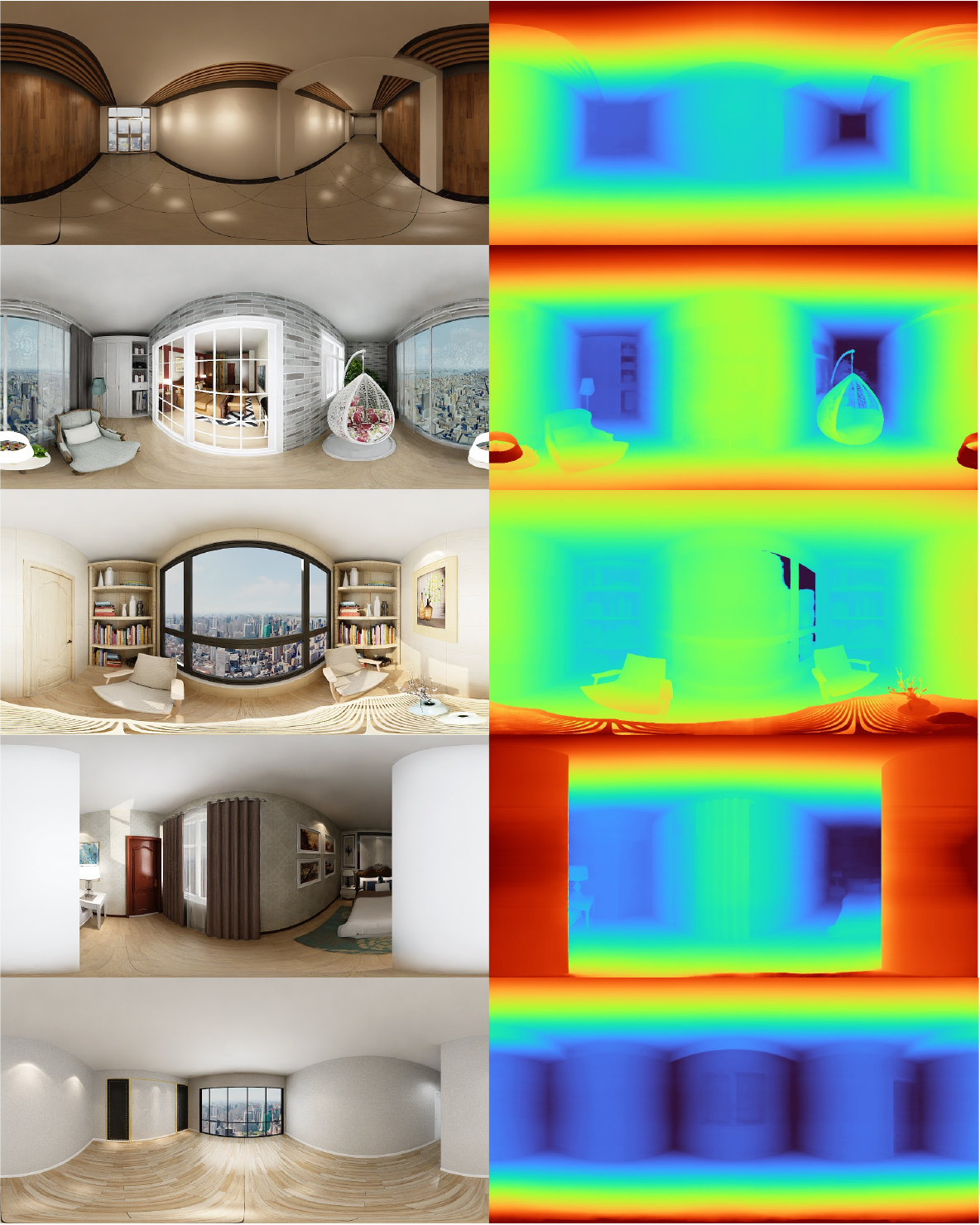}
    \caption{\textbf{Generalization to Unseen Visual Domains.} Visualizing depth predictions on \textbf{panoramic images}, an unseen domain during training. Our model successfully handles such extreme distortion and novel viewpoints.}
\label{fig:supp_vis_panorama}
\end{figure}

\begin{figure}[h]
    \centering
    \includegraphics[width=\linewidth]{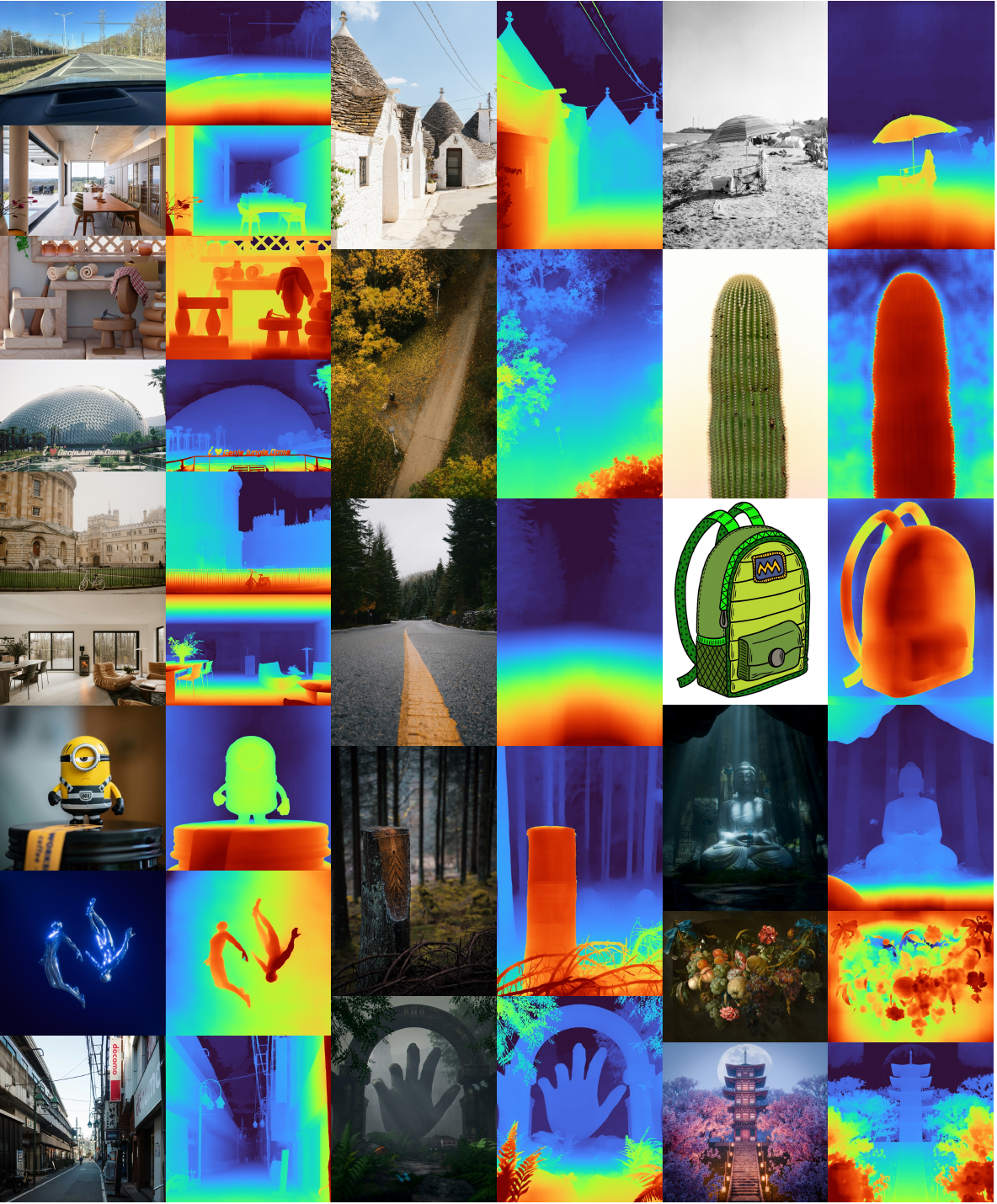}
    \caption{\textbf{Generalization to Unseen Visual Domains.} Additional visualizations of depth predictions on diverse in-the-wild images.}
\label{fig:supp_vis_anything1}
\end{figure}

This subsection evaluates the model's zero-shot generalization on monocular depth estimation across visual domains absent from training. Tests were conducted without prompt guidance on three challenging scenarios: 
\textbf{panoramic images} from spherical projections, \textbf{fisheye images} with extreme distortions, and diverse \textbf{in-the-wild scenes} including cartoons, grayscale images, and artistic renderings.
Qualitative results (Fig.~\ref{fig:supp_vis_fisheye},  Fig.~\ref{fig:supp_vis_anything2}, Fig.~\ref{fig:supp_vis_panorama}, and Fig.~\ref{fig:supp_vis_anything1}) confirm accurate metric depth estimation throughout. This robust performance across domains previously unrepresented in training data substantiates our claim of achieving “Metric Anything” generalization.

\section{Training Details}


\subsection{Training and Test Set Split.} Across all experiments, results are reported in a zero-shot manner, meaning that the training and test sets are sourced from completely distinct datasets with no shared origin. The only exception is ScanNet \cite{dai2017scannet}, where we utilize its training set for model training and evaluate on its test set in the multi-view 3D metric reconstruction experiments, following the exact protocol of Map Anything \cite{keetha2025mapanything}. This usage of ScanNet for training is fair to other methods, as it is also used for training in Map Anything. 

\subsection{Training Details of Pre-trained Model.} We train our model on 144 H200 GPUs for 100k steps, using a 10k-step warm-up, a peak learning rate of 1$\times$10$^{-6}$ for the ViT backbone, and 1$\times$10$^{-5}$ for the DPT head and the prompt layer. To incorporate the prompt, we introduce the required convolutional kernels and biases with zero initialization.

\subsection{Training Details of Distilled Model.}
We distill the pre-trained model into a prompt-free student, supervised by pseudo-labels generated through pre-trained model inference. To validate the generality of our pre-trained model, we design two training paradigms: (1) From-scratch training, where we initialize the ViT backbone from DINOv3 ViT-H+/16 \cite{simeoni2025dinov3} and randomly initialize the DPT head. We train our model on 144 H200 GPUs for 200k steps, using a 5k-step warm-up with a peak learning rate of 2$\times$10$^{-6}$ for the ViT backbone and 2$\times$10$^{-5}$ for the DPT head. (2) Fine-tuning state-of-the-art methods: We initialize all network parameters from MoGo-2 \cite{wang2025moge2}. We train our model on 80 H200 GPUs for 10k steps, employing a 1k-step warm-up with a peak learning rate of 2.5$\times$10$^{-6}$ for the ViT backbone and 2.5$\times$10$^{-5}$ for the DPT head.

For all experiments mentioned above, including pre-training and distillation, we supervise the network using the original resolution of depth maps from the datasets. Specifically, inputs are uniformly resized to $1536 \times 1536$, and the network outputs are produced at the same resolution. However, for loss computation during training supervision, we resize the network outputs to match the original dataset resolution for comparison with ground truth. To enhance training efficiency, we utilize FlashAttention \cite{dao2022flashattention} alongside DeepSpeed with ZeRO Stage-2 optimizer \cite{rasley2020deepspeed} and BF16 precision.

\subsection{Training Details of Vision-Language-Model.}
Our VLM is initialized from LLaVa-NeXT-Video-7B \cite{zhang2024llavanext-video} and equipped with two visual encoders, a SigLIP based image encoder and a our pretrained ViT of Metric Anything. The  features of Metric Anything are first mapped to a 1152 dimensional space using a lightweight adapter. We then fuse the SigLIP features and the adapted Metric Anything  features with a cross attention module, and apply a projection layer to match the multimodal embedding dimension of the backbone before feeding the fused representation into the language model. The projection layer is initialized from the pretrained LLaVA-NeXT model, while the adapter and cross attention module are initialized from scratch.
For training, we adopt the dataset configuration introduced in VLM-3R \cite{bhat20253d} and fine-tune our VLM for 2K steps with a per-device batch size of 4, using 10 nodes with 8 GPUs per node. During training, the LLM backbone, the vision adapter, the cross-attention fusion modules, and the projection layer are all kept trainable, while all remaining vision encoders are frozen. We use AdamW with a learning rate of 1e-5 for the trainable modules, a cosine learning rate schedule with a 0.03 warmup ratio, and gradient clipping with a threshold of 0.5.

\subsection{Training Details of Vision-Language-Action Model.}
For each frame, we first extract a depth map using our  distilled depth estimation model. The depth maps are then normalized using the 5th and 95th percentiles and subsequently converted into depth tokens using a VQ-VAE, which are used as an extra ground-truth signal to supervise the VLA model in predicting depth tokens, while simultaneously predicting the action token..
We apply parameter-efficient fine-tuning via LoRA, attaching low-rank adaptation modules to all Transformer layers in the model. The model is trained on 10 nodes with 8 GPUs per node, using a global batch size of 160. We use AdamW with a learning rate of 5e-4 and a cosine learning rate schedule. 

\paragraph{Evaluation Metrics}
\paragraph{Depth Metrics.}
For the quantitative evaluation of depth estimation, we follow the standard metrics from prior works. Let $d_i$ and $\hat{d}_i$ denote the ground truth depth and the predicted depth for pixel $i$, respectively. The evaluation is performed over all $N$ valid pixels. Our reported metrics are defined as follows:

\begin{itemize}
    \item \textbf{Threshold Accuracy ($\delta_i$):} The percentage of pixels where the ratio of predicted and ground truth depth falls within a certain threshold:
    \[
        \% \text{ of pixels s.t. } \max\left(\frac{\hat{d}_i}{d_i}, \frac{d_i}{\hat{d}_i}\right) < \delta, 
    \]
    where $\delta \in \{1.25, 1.25^2, 1.25^3\}$.

    \item \textbf{Absolute Relative Error (AbsRel):}
    \[
        \frac{1}{N} \sum_{i} \frac{|\hat{d}_i - d_i|}{d_i}
    \]


    \item \textbf{Root Mean Squared Error (RMSE):}
    \[
        \sqrt{\frac{1}{N} \sum_{i} (\hat{d}_i - d_i)^2}
    \]
    
    \item \textbf{Mean Absolute Error (MAE):}
    \[
        \frac{1}{N} \sum_{i} |\hat{d}_i - d_i|
    \]


    \item \textbf{Log10 Error:}
    \[
        \frac{1}{N} \sum_i |\log_{10}(\hat{d}_i) - \log_{10}(d_i)|
    \]
    
    
\end{itemize}

These metrics are used throughout our experiments. Specifically:
\begin{itemize}
    \item \textbf{$\delta_1$}: Reported in Sec~\ref{sec:experiment} (Tab.~\ref{tab:student_mde_vs}, Tab.~\ref{tab:improve_mapanything}) for zero-shot monocular depth estimation evaluation.
    \item {AbsRel}: Reported in Sec~\ref{sec:experiment} (Tab.~\ref{tab:zeroshot_completion}, Tab.~\ref{tab:improve_mapanything}) for metric depth accuracy evaluation.
    \item {RMSE, MAE}: Reported in Sec~\ref{sec:experiment} (Tab.~\ref{tab:radar-comparison}) for radar-camera depth estimation evaluation.
    \item Log${_{10}}$, ${\delta}_{2}$, ${\delta_3}$: Used in comprehensive evaluation tables in the appendix for additional depth accuracy assessment.
\end{itemize}

\paragraph{Boundary Metrics.}
Following DepthPro~\cite{depthpro}, we evaluate boundary sharpness using depth-based and mask-based metrics. For depth maps, we define occluding contours based on pairwise depth ratios between neighboring pixels. Let $i,j$ be the locations of two neighboring pixels. We define an occluding contour $c_d$ derived from a depth map $d$ as:
\[
c_d(i,j) = \left[ \frac{d(j)}{d(i)} > \left(1 + \frac{t}{100}\right) \right],
\]
where $[\cdot]$ is the Iverson bracket, indicating the presence of an occluding contour if the depth differs by more than $t\%$.
For all pairs of neighboring pixels, we compute precision ($P$) and recall ($R$) as:
\[
    \text{P}(t) = \frac{\sum_{i,j\in N(i)} c_d(i,j) \wedge c_{\hat{d}}(i,j)}{ \sum_{i,j\in N(i)} c_{d}(i,j)},
\]
\[
    \text{R}(t) = \frac{\sum_{i,j\in N(i)} c_d(i,j) \wedge c_{\hat{d}}(i,j)}{ \sum_{i,j\in N(i)} c_{\hat{d}}(i,j)},
\]
where $N(i)$ denotes the set of neighboring pixels of $i$, $c_d$ and $c_{\hat{d}}$ are occluding contours from predicted and ground-truth depth maps, respectively.
The \textbf{Boundary F1 score} is computed as:
\begin{align}
\text{F1}(t) = \frac{2 \cdot \text{P}(t) \cdot \text{R}(t)}{\text{P}(t) + \text{R}(t)}.
\end{align}

\paragraph{Camera Intrinsics Metrics.}
For focal length estimation evaluation, we report the mean and median angular errors in degrees. Given the ground-truth focal length $f_{\text{gt}}$ (in pixels) and image width $w$, the horizontal field of view (FOV) is computed as:
\[
\text{FOV} = 2 \arctan\left(\frac{w}{2 f_{\text{gt}}}\right).
\]

The angular error in degrees is:
\[
\text{Error} = |\text{FOV}_{\text{pred}} - \text{FOV}_{\text{gt}}| \times \frac{180}{\pi},
\]
where $\text{FOV}_{\text{pred}}$ and $\text{FOV}_{\text{gt}}$ are predicted and ground-truth FOV values, respectively.
FOV estimation results are reported in Sec~\ref{sec:experiment} (Tab.~\ref{tab:fov-comparison}) for camera calibration evaluation.

\subsection{Loss Functions}
\label{sec:loss_functions}


\paragraph{Pre-train Model Losses.}
The pre-train model is trained to predict dense metric depth from monocular images conditioned on sparse metric prompts. We operate in the inverse depth space $C$. 
For real-world metric datasets, we adopt a \textbf{Robust MAE} loss that discards the top $20\%$ of pixels with the largest errors per image to mitigate the influence of noisy ground truth:
\begin{equation} \label{eq:robust_mae}
\mathcal{L}_{\mathit{MAE}}(\hat{C}, C) = \frac{1}{N'} \sum_{i \in \mathcal{S}} |\hat{C}_i - C_i|,
\end{equation}
where $\mathcal{S}$ is the set of pixels after removing the top $20\%$ largest errors, and $N' = |\mathcal{S}|$.
For synthetic datasets without metric scale, we additionally apply the \textbf{Scale-and-Shift-Invariant Mean Absolute Gradient Error (SSI-MAGE)} loss~\cite{depthpro}. First, we normalize predictions and ground truth via mean absolute deviation from the median to achieve scale and shift invariance:
\begin{align}
\tilde{C} &= \frac{C - \text{median}(C)}{\text{MAD}(C)}, \\
\tilde{\hat{C}} &= \frac{\hat{C} - \text{median}(\hat{C})}{\text{MAD}(\hat{C})},
\end{align}
where $\text{MAD}(C) = \text{median}(|C - \text{median}(C)|)$ is the median absolute deviation.
Then, we compute the multi-scale gradient loss. Let $\nabla_S$ denote the Scharr gradient operator~\cite{Scharr1997}. The multi-scale derivative loss over $M$ scales is defined as:
\begin{equation} \label{eq:multiscale_gradient}
\mathcal{L}_{*, p, M}(C, \hat{C}) = \frac{1}{M}\sum^M_j\frac{1}{N_j}\sum_i^{N_j} |\nabla_{*} C^j_i - \nabla_{*} \hat{C}^j_i|^p,
\end{equation}
where $C^j$ and $\hat{C}^j$ are the inverse depth maps at scale $j$ (obtained by blurring and downsampling by a factor of 2 per scale), and $N_j$ represents the number of valid pixels at scale $j$.
The \textbf{SSI-MAGE} loss is defined as:
\begin{equation} \label{eq:ssi_mage}
\mathcal{L}_{\mathit{SSI\text{-}MAGE}}(\hat{C}, C) = \mathcal{L}_{S,1,6}(\tilde{\hat{C}}, \tilde{C}),
\end{equation}
where $\mathcal{L}_{S,1,6}$ denotes the multi-scale gradient loss with Scharr operator ($S$), $L_1$ norm ($p=1$), and 6 scales ($M=6$).
The overall training objective for the teacher model is:
\begin{equation} \label{eq:teacher_loss}
\mathcal{L}_{\mathit{total}} = \alpha \mathcal{L}_{\mathit{MAE}} + \beta \mathcal{L}_{\mathit{SSI\text{-}MAGE}},
\end{equation}
where $\alpha$ and $\beta$ are weighting coefficients, we set $\alpha=15$ and $\beta=5$. For synthetic datasets, both terms are applied; for real-world datasets, only the robust MAE loss is used. This loss formulation is described in Sec~\ref{sec:pretrain} of the main manuscript.

\paragraph{Distill Model Losses.}
The distill model is trained on pseudo labels generated by the teacher model. To handle the wide depth range (from near to far distances) in the pseudo labels (as shown in Fig.~\ref{fig:depthrange}), we design a \textbf{Distance-Balanced Inverse-Depth Loss}.
The loss operates in a log-space representation that preserves fine-grained sensitivity in near regions while extending effective supervision to long-distance areas. The transformed depth value is defined as (Eq.~\ref{eq:ours_log_distance} in the main manuscript):
$D_{\log} = 1 - {\ln(D)}/{\ln(C)},$
where $D$ is the metric depth and $C=400$ is a hyperparameter controlling the trade-off between long-range and short-range supervision.
The distance-balanced loss is then computed as:
\begin{equation} \label{eq:student_loss}
\mathcal{L}_{\mathit{Student}}(\hat{D}, D) = \frac{1}{N}\sum^N_i |D_{\log}(\hat{D}_i) - D_{\log}(D_i)|,
\end{equation}
where $\hat{D}$ and $D$ are predicted and ground-truth metric depth maps, respectively.
This loss function addresses the limitation of standard inverse-depth loss, which decays too rapidly with distance, by providing more balanced supervision across the full depth range covered by our teacher-generated pseudo labels. Additionally, we apply the \textbf{SSI-MAGE} loss (Eq.~\ref{eq:ssi_mage}) to the log-space transformed depth maps, combining it with our distance-balanced supervision to further enhance boundary sharpness and geometric detail preservation. The overall student training objective is:
\begin{equation} \label{eq:student_total_loss}
\mathcal{L}_{\mathit{StudentTotal}} = \gamma \mathcal{L}_{\mathit{Student}} + \delta \mathcal{L}_{\mathit{SSI\text{-}MAGE}}(D_{\log}(\hat{D}), D_{\log}(D)),
\end{equation}
where $\gamma$ and $\delta$ are weighting coefficients, we set $\gamma=10$ and $\delta=2$. The design rationale and experimental validation are presented in Sec~\ref{sec:student} of the main paper and Tab.~\ref{tab:training_objectives}.
\FloatBarrier

\section{Limitations}

Our work maintains the central projection camera assumption and has not been extended to specialized camera models (e.g., non-central or non-pinhole configurations). In terms of model scalability, while our data-centric scaling strategy demonstrates strong empirical gains, the scalability of the model architecture itself remains unexplored. Expanding the model's architecture could potentially enhance its capability for depth perception in more complex and diverse scenarios.

\section{Conclusion}

We present \textbf{Metric Anything}, a scalable pretraining framework for metric depth estimation that learns from diverse, noisy 3D sources without task-specific architectures or manually engineered prompts. Using Sparse Metric Prompts to separate spatial reasoning from sensor and camera biases, our approach effectively leverages heterogeneous data. Experiments reveal, for the first time, a clear scaling effect in the metric depth trick. Both the pretrained model and its distilled prompt-free student achieve state-of-the-art results across a wide range of downstream tasks. These results indicate more efficient general-purpose solutions for real-world depth perception.

{
\bibliographystyle{plain}
\bibliography{main}
}

\end{document}